%% file: RealSR-R1.tex
\documentclass[10pt,journal,compsoc]{IEEEtran}
\usepackage{times}
\usepackage{epsfig}
\usepackage{graphicx}
\usepackage{amsmath}
\usepackage{amssymb}
\usepackage{algorithm}
\usepackage{algorithmic}
\usepackage{graphics}
\usepackage{epstopdf}
\usepackage{subfigure}
\usepackage{multirow}
\usepackage{url}
\usepackage{mathtools}
\usepackage{amsthm}
\usepackage{makecell}
\usepackage{orcidlink}
\usepackage{wrapfig}
\usepackage{caption}
\usepackage{adjustbox}
\usepackage{booktabs}
\ifCLASSOPTIONcompsoc
  \usepackage[nocompress]{cite}
\else
  \usepackage{cite}
\fi

\begin{document}
\title{RealSR-R1: Reinforcement Learning for Real-World Image Super-Resolution with Vision-Language Chain-of-Thought}
\author{Junbo Qiao{$\dag$}, Miaomiao Cai{$\dag$}, Wei Li, Xudong Huang, Jie Hu, Xinghao Chen, Shaohui Lin{$^*$}, Hongkai Xiong~\IEEEmembership{Fellow,~IEEE}% <-this % stops a space
\IEEEcompsocitemizethanks{\IEEEcompsocthanksitem J. Qiao and S. Lin are with the School of Computer Science and Technology, East China Normal University,
Shanghai, 200062, China. \protect\\
E-mail: shlin@cs.ecnu.edu.cn
\IEEEcompsocthanksitem M. Cai is with the University of Science and Technology of China,
Hefei, China.
\IEEEcompsocthanksitem W. Li, X. Huang, J. Hu and X. Chen are with the Huawei Noah's Ark Lab, Huawei Technologies Ltd,
Shanghai, China.
\IEEEcompsocthanksitem Hongkai Xiong is with the East China Normal University,
Shanghai, 200062, China, and the Shanghai Jiaotong University, Shanghai, China.
\IEEEcompsocthanksitem (*Corresponding author: Shaohui Lin)
\IEEEcompsocthanksitem ($^{\dag}$These authors contributed equally to this work.)
}}

\markboth{SUBMISSION TO IEEE TRANSACTIONS ON PATTERN ANALYSIS AND MACHINE INTELLIGENCE}%
{Shell \MakeLowercase{\textit{et al.}}: Bare Demo of IEEEtran.cls for Computer Society Journals}

\IEEEtitleabstractindextext{
\begin{abstract}
Real-World Image Super-Resolution is one of the most challenging task in image restoration. However, existing methods struggle with an accurate understanding of degraded image content, leading to reconstructed results that are both low-fidelity and unnatural. We present RealSR-R1 in this work, which empowers the RealSR models with understanding and reasoning capabilities. Inspired by the success of Chain of Thought (CoT) in large language models (LLMs), we simulate the human process of handling degraded images and propose the VLCoT framework, which integrates vision and language reasoning. The framework aims to precisely restore image details by progressively generating more comprehensive text and higher-resolution images. To overcome the challenge of traditional supervised learning CoT failing to generalize to real-world scenarios, we introduce, for the first time, Group Relative Policy Optimization (GRPO) into the Real-World Image Super-Resolution task. We propose VLCoT-GRPO as a solution, which designs four reward functions: (1) Format reward, used to standardize the CoT process; (2) Degradation reward, to incentivize accurate degradation estimation; (3) Understanding reward, to ensure the accuracy of the generated content; and (4) Generation reward, where we propose using a visual expert model to evaluate the quality of generated images, encouraging the model to generate more realistic images. Extensive experiments demonstrate that our proposed RealSR-R1 can generate realistic details and accurately understand image content, particularly in semantically rich scenes or images with severe degradation. Our
source codes have been released at \href{https://github.com/Junboooo/RealSR-R1}{https://github.com/Junboooo/RealSR-R1}
\end{abstract}

}

\maketitle

\IEEEdisplaynontitleabstractindextext

% \fi
\IEEEpeerreviewmaketitle

\begin{figure*}[t]
    \centering
    \includegraphics[width=0.99\textwidth]{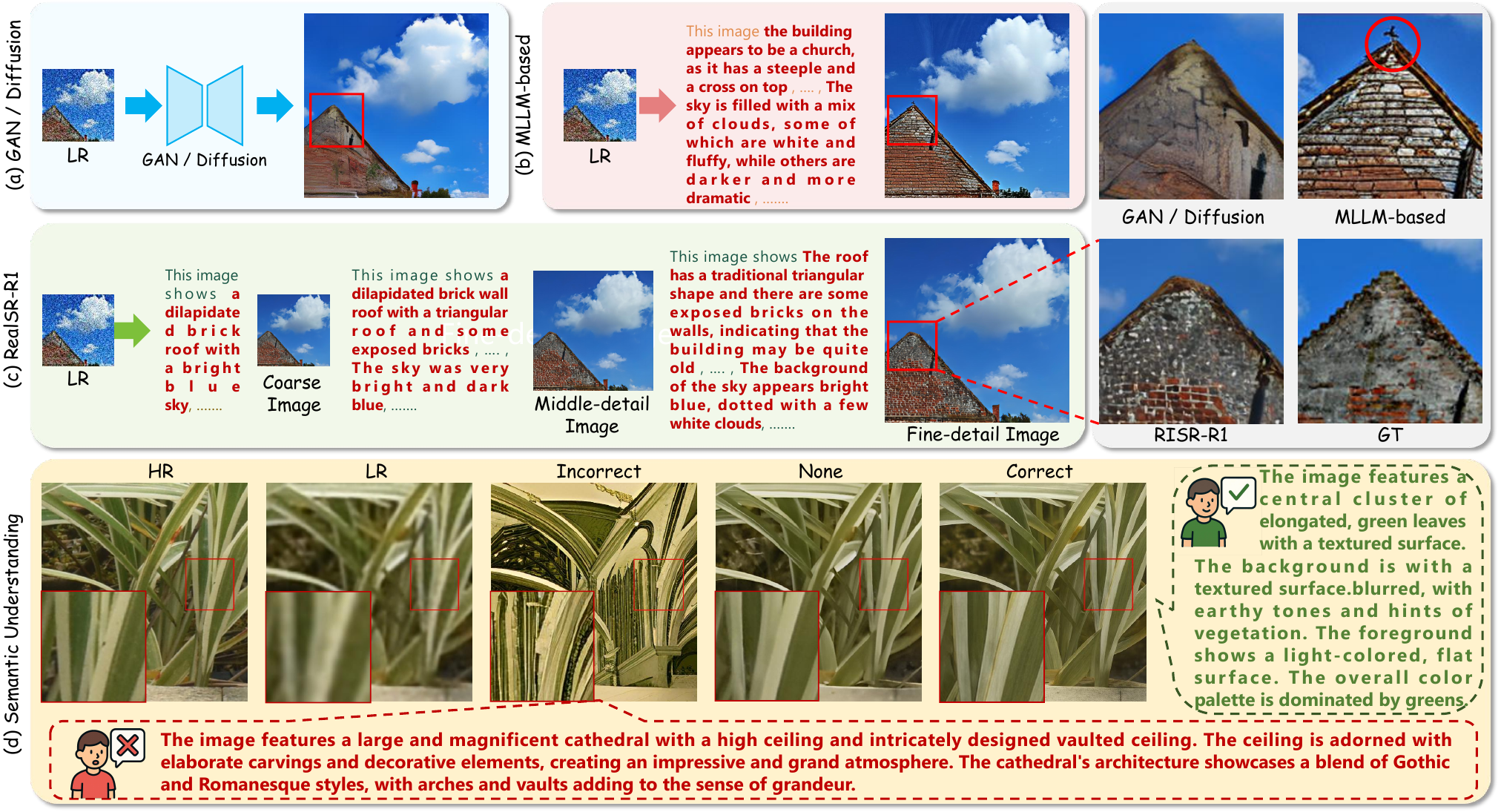} 
    %\vspace{-2mm}
    \caption{(a) The GAN and diffusion methods themselves do not possess the capability to understand the content of images. (b) Directly interpreting and generating detailed descriptions of degraded images leads to incorrect understanding and image restoration. (c) RealSR-R1 simulates the human image restoration process by progressively refining the understanding of image content and generating higher-quality images. (d) Incorrect semantics lead the model to reconstruct erroneous content, whereas correct semantic recovery of fine-grained details, such as plant textures, is crucial.}
    \label{fig:teaser}
    \vspace{-4mm}
\end{figure*}

\IEEEraisesectionheading{\section{Introduction}\label{intro}}

\IEEEPARstart{R}{eal}-world image super-resolution (RealSR) aims to reconstruct perceptually realistic high-quality (HQ) images from low-quality (LR) inputs that suffer from real-world degradations. Due to the complexity of these degradations, traditional methods~\cite{dong2014learning,liang2021swinir,lim2017enhanced,zhang2018image} constrained by $L_{1}$ or $L_{2}$ losses may produce overly smooth outcomes, prompting many researchers to introduce GAN-based~\cite{ledig2017photo,wang2021real} and diffusion-based~\cite{zhang2023adding,yang2024pixel,yu2024scaling,wu2024seesr,chen2024adversarial,sun2024pixel,wu2024one,cheng2025effective} methods into RealSR tasks with remarkable success. 
However, when handling severely degraded inputs, these methods have limited ability to understand low-level image content. 
As shown in Fig.~\ref{fig:teaser}(a), they struggle to understand heavily degraded images (\textit{e.g.}, the roof of a dilapidated brick wall), often resulting in the generation of unnatural or implausible objects.

Recently, Multimodal Large Language Model (MLLM) leverages high-capacity networks and large-scale data training, demonstrating preliminary image understanding and generation capabilities in RealSR tasks~\cite{wei2025perceive}. 
%Despite the stronger understanding capabilities of such methods, directly providing detailed descriptions of severely degraded and semantically ambiguous scenarios leads to content misinterpretation and detail confusion, resulting in artifacts, random generation, or structurally inconsistent outputs. 
However, we observe that in Fig.~\ref{fig:teaser}(b), the severely degraded LR depicts a scene of rooftops and sky, yet the MLLM misinterprets this scene as "the building appears to be a church, as it has a steeple and a cross on top," consequently producing hallucinated structures in the reconstructed image. This phenomenon exposes a fundamental issue in directly applying the semantics-driven generation logic of MLLMs to image reconstruction tasks: when confronted with heavily degraded LR inputs, the model may fail to correctly interpret the visual content and instead rely on its internal, generalized semantic priors to “hallucinate” plausible but incorrect details.

To further validate the impact of semantic understanding correctness on reconstruction quality, we design a controlled experiment based on the PURE~\cite{wei2025perceive} by explicitly manipulating the correctness of semantic understanding (incorrect, none, and correct). As shown in Fig.~\ref{fig:teaser} (d), incorrect semantic understanding leads the model to generate erroneous structural content, whereas only accurate semantic interpretation enables the recovery of fine-grained details such as plant textures. These observations raise a key question: \textit{how can a model acquire accurate semantic understanding when the input is severely degraded?}

For this question, we first reflect on how humans perform visual restoration. We observe that semantic understanding and image reconstruction are not independent processes but are inherently interdependent~\cite{wang2025autoregressive,liu2024semantics,jiang2025t2i}. Specifically, humans typically begin by assessing the degree of degradation and forming a coarse semantic understanding of the scene, rather than attempting pixel-level restoration directly. The reconstruction process then proceeds in a coarse-to-fine manner, during which semantic understanding is progressively refined as more visual details are recovered. This iterative interaction ultimately yields both a visually detailed reconstruction and an accurate semantic interpretation. Therefore, establishing a bidirectional reasoning loop between visual restoration and semantic understanding is crucial.

Motivated by these insights, we propose the vision-language Chain-of-Thought (\textbf{VLCoT}), a multi-step vision-language reasoning framework that models the mutual reinforcement between semantic understanding and image reconstruction. Instead of performing one-shot restoration based on uncertain semantics, VLCoT adopts a coarse-to-fine reasoning paradigm, where vision and language iteratively refine each other. As shown in Fig.~\ref{fig:teaser}(c), given a degraded LR image, the model first produces an initial coarse understanding along with a coarse reconstructed image. These intermediate results then serve as improved visual and semantic cues for subsequent reasoning steps, progressively enhancing both the textual understanding and image fidelity. Ultimately, VLCoT achieves accurate recognition of fine-grained structures—such as the brick walls and roof—and produces the most faithful high-resolution restoration.

%Motivated by these insights, we propose \textbf{VLCoT}, a multi-step reasoning process that combines vision and language in a coarse-to-fine manner. As shown in Fig.~\ref{fig:teaser}(c), for the input LR, the model first estimates the degradation degree, coarse understanding, and coarse restored image, and then progressively generates more detailed textual understanding descriptions and higher-resolution images. Ultimately, the proposed VLCoT accurately understands the brick walls and roofs in the image and achieves the most precise image restoration.

%Furthermore, we propose \textbf{VLCoT-GRPO}, which leverages Group Relative Policy Optimization (GRPO)~\cite{shao2024deepseekmath} to enhance and optimize the image-text integrated reasoning process of VLCoT.
%We introduce reinforcement learning (RL) for two reasons: (1) Recent MLLMs have already demonstrated basic CoT reasoning capabilities in individual modalities such as text or image. To address this limitation, we aim to foster active exploration that unlocks synergistic reasoning across modalities. 
%(2) Supervised fine-tuning (SFT) relies on imitating high-quality reasoning paths, which may lead the model to adopt "seemingly correct" patterns while failing to identify and differentiate flawed reasoning effectively.
%In contrast, RL encourages exploration of diverse reasoning paths, strengthens correct reasoning patterns through reward signals, and simultaneously identifies and suppresses incorrect reasoning patterns. This approach is better suited for non-regular problems like RealSR.

While VLCoT establishes a mutual interaction between semantic understanding and visual reconstruction, its reasoning quality still depends heavily on supervised learning, which limits the model’s ability to self-correct and generalize.
To further enhance this process, we propose \textbf{VLCoT-GRPO}, which adapts Group Relative Policy Optimization (GRPO)~\cite{shao2024deepseekmath} to optimize the image-text integrated reasoning of VLCoT through reinforcement learning (RL).
We introduce RL for two main reasons.
(1) Although recent MLLMs exhibit basic chain-of-thought (CoT) reasoning within individual modalities (e.g., text or image), they struggle to perform synergistic reasoning across modalities. To address this limitation, we aim to foster active exploration that unlocks synergistic reasoning across modalities. 
(2) Supervised fine-tuning (SFT) relies on imitating high-quality reasoning paths, and this approach risks overfitting to “seemingly correct” but suboptimal patterns; in contrast, RL provides reward-driven feedback that strengthens genuinely correct reasoning behaviors while penalizing erroneous ones. This approach is better suited for non-regular problems like RealSR.

%Consequently, VLCoT-GRPO promotes diverse reasoning exploration, reinforces robust semantic-visual alignment, and further improves the model’s capability to reason coherently and reconstruct faithfully in the challenging RealSR setting.

%Unlike standard NLP tasks that typically employ clear-cut reward signals, RealSR lacks a unified, standardized reward function, posing challenges for RL-based optimization. To tackle these issues, we propose four reward functions to comprehensively guide the VLCoT generation process: (1) Format Reward: Enforces structural conventions to ensure a well-formatted, hierarchical generation process. (2) Degradation Reward: Ensure accurate perception of degradation. (3) Understand Reward: Assesses the model’s accuracy in understanding the LR image contents, thus ensuring reasonableness and precision of the outputs. (4) Generation Reward: Using a powerful prior visual expert model, a multi-faceted evaluation of the image quality, realism and consistency of a group of outputs is provided.

Unlike standard NLP or reasoning tasks where explicit reward signals (e.g., accuracy or token-level correctness) are readily available, RealSR lacks a unified and standardized reward metric, making RL-based optimization particularly challenging. The quality of restoration involves both semantic fidelity and visual realism, which cannot be captured by a single scalar reward. 
To address this issue, we design four complementary reward functions to comprehensively guide the vision-language reasoning process of VLCoT-GRPO:
(1) Format Reward: enforces structural and hierarchical generation conventions, ensuring a consistent reasoning format;
(2) Degradation Reward: evaluates whether the model correctly perceives the degradation level, which is critical for guiding restoration difficulty;
(3) Understanding Reward: measures the semantic accuracy of the model’s interpretation of LR content, ensuring logical and perceptual coherence; and
(4) Generation Reward: employs a powerful pretrained visual expert to provide multi-dimensional assessments of image realism, quality, and consistency across outputs.
These complementary rewards establish a holistic feedback mechanism that drives VLCoT-GRPO to achieve more stable training, stronger semantic-visual alignment, and superior reconstruction performance under severe degradations.

Building on the proposed vision-language reasoning framework and its reinforcement optimization strategy, we propose \textbf{RealSR-R1}, the first reasoning-enhanced RealSR model. Benefiting from the multi-level reward guidance and cross-modal reasoning refinement, RealSR-R1 not only achieves superior quantitative performance but also produces perceptually realistic results that align more closely with human visual preferences. Extensive experiments further demonstrate its robustness in handling challenging degradations and semantically intricate scenes, marking a significant step toward reasoning-aware RealSR.

The main contributions of this paper can be summarized as follows:
\begin{itemize}
\item We propose \textbf{RealSR-R1}, the first reasoning-enhanced solution to the RealSR problem, highlighting the importance of correct semantic understanding for reconstruction to achieve more stable and trustworthy image reconstruction.
\item We propose \textbf{VLCoT}, a progressive multi-step vision-language reasoning framework that emulates the human process of “understanding before reconstructing.” VLCoT jointly refines semantic understanding and visual restoration in a coarse-to-fine manner.
\item We present \textbf{VLCoT-GRPO}, a reinforcement learning framework for multi-modal joint reasoning, which aligns and optimizes visual and textual reasoning with multiple reward signals, enhancing semantic-visual coherence and reasoning robustness.
\end{itemize}

\begin{figure*}[t]
    \centering
    \includegraphics[width=0.99\textwidth]{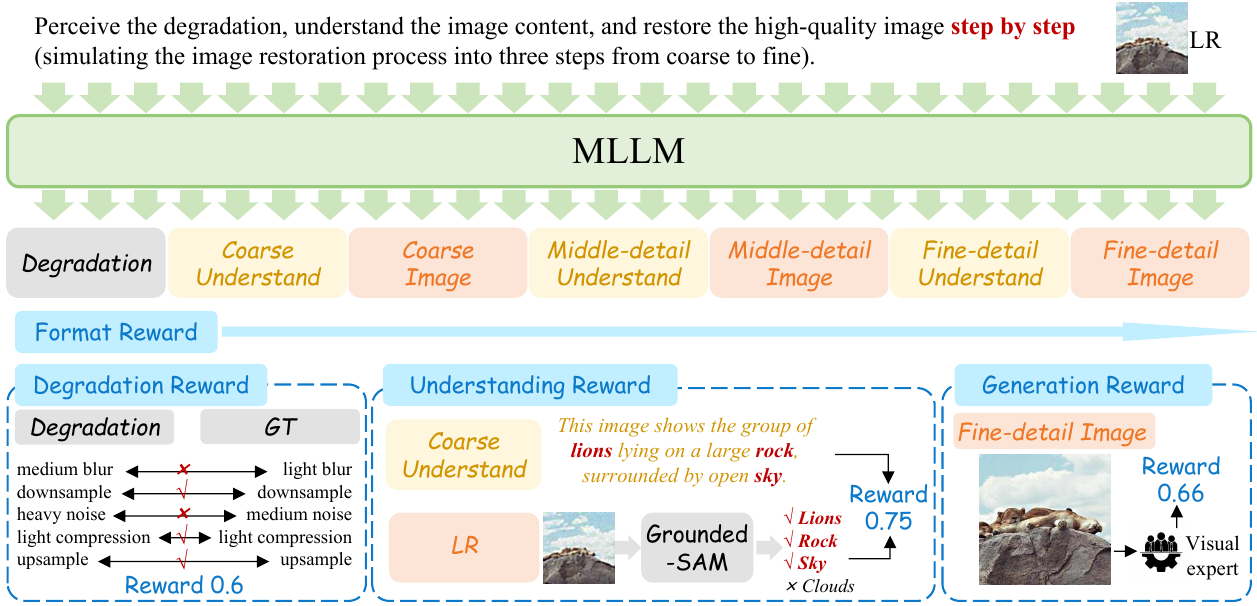} 
    %\vspace{-2mm}
    \caption{llustration of the proposed RealSR-R1. The multi-step output in the center of the image represents the VLCoT process. Four specially designed reward functions are displayed at the bottom, including format reward, degradation reward, understanding reward, and generation reward.}
    \label{fig:method}
    \vspace{-0.1cm}
\end{figure*}

\section{Related Work}
\label{RW}
\subsection{Real-World Image Super-Resolution.}
Classical image restoration methods~\cite{liang2021swinir,lim2017enhanced,zhang2018image, qiao2025lipt,qiao2024hi} trained with L1 or L2 loss struggle to generate high-frequency details in the Real SR task, often resulting in overly smooth outputs. Researchers have attempted to introduce GANs~\cite{ledig2017photo,wang2021real,qiao2025dcs} into this field, but the training process is unstable, and the lack of understanding of the image content also leads to the generation of artifacts. Diffusion models have gained widespread popularity due to their strong generative capabilities. Some recent methods leverage pre-trained diffusion backbones (e.g., SD~\cite{sd} or FLUX~\cite{flux2024}) and employ ControlNet~\cite{zhang2023adding} and LR inputs to steer the generation process. 
Other methods enhance semantic alignment by leveraging additional text understanding models to extract textual descriptions~\cite{yang2024pixel} or class labels~\cite{wu2024seesr} from the LR input. One-step diffusion methods~\cite{wu2024one,sun2024pixel,chen2024adversarial,he2024one,wu2025one} direct mapping from LR to HR images with additional LoRA~\cite{hu2022lora}. Despite notable improvements in perceptual quality, these approaches still struggle in complex scenarios due to limited semantic understanding, resulting in low-fidelity content representation and unnatural local structures. Recently, Wei et al.~\cite{wei2025perceive} introduced MLLMs into the RealSR task, significantly enhancing the model's semantic understanding. However, unlike natural images, RealSR inputs typically suffer from severe degradation, making it challenging to directly obtain detailed and accurate image descriptions. In this paper, we propose VLCoT, a vision-language CoT framework to simulate the process of human image restoration, progressively generating detailed image content descriptions and recovering high-frequency details.

\subsection{Chain-of-Thought.}
In natural language processing (NLP) and MLLM, researchers have observed that directly mapping inputs to outputs on complex tasks often leads to biased comprehension or suboptimal outcomes~\cite{wei2022chain}. Consequently, the Chain-of-Thought (CoT) mechanism~\cite{wei2022chain,kojima2022large,jiang2025mme,zhang2024mathverse} is proposed, guiding models to simulate the human thinking process by explicitly constructing intermediate reasoning steps. CoT significantly enhances model understanding and output reliability~\cite{yao2023tree,mondal2024kam,mu2023embodiedgpt,zhang2023multimodal}. However, existing CoT approaches in vision and vision-language domains typically perform reasoning within a single modality, relying solely on textual descriptions~\cite{shao2024visual,liu2025visual,pan2025medvlm} or visual features~\cite{guo2025can}. Such designs are insufficient for Real Image Super-Resolution (RealSR), where both semantic understanding and visual reconstruction are inherently intertwined. 
Motivated by the mutual reinforcement between understanding and reconstruction, we propose the Vision-Language Chain-of-Thought (VLCoT) framework tailored for RealSR. VLCoT adopts a coarse-to-fine, alternating reasoning paradigm, where semantic reasoning provides guidance for visual restoration, and intermediate reconstruction results in turn refine semantic understanding. This bidirectional reasoning process enables the model to progressively align semantics and perception, achieving coherent and faithful image reconstruction even under severe degradations.

\subsection{Reinforcement Learning.}
Reinforcement learning (RL) has recently gained significant attention for its effectiveness in enhancing reasoning capabilities of large models. The o1~\cite{jaech2024openai} model proposed by OpenAI has attracted widespread attention, particularly for its advancements in enhancing reasoning capabilities through RL. DeepSeek-R1~\cite{guo2025deepseek} introduced a rule-based reward mechanism, Group Relative Policy Optimization (GRPO), a training paradigm that requires comprehensive reasoning before final answer generation. In recent years, this methodology has been widely adopted in the development of MLLMs~\cite{chenvinci,meng2025mm,yang2025r1,zhang2025r1,deng2025openvlthinker} with task-specific reward functions, such as correctness~\cite{liu2025visual} or IoU~\cite{liu2025seg}. 
However, most of these tasks primarily apply reinforcement learning to pure text~\cite{shao2024visual,liu2025visual,pan2025medvlm} or image~\cite{guo2025can,cai2025dspo} inference processes. In contrast, RealSR requires joint optimization of semantic reasoning and visual reconstruction, which are tightly coupled yet rarely addressed under the RL framework.
To bridge this gap, we explore RL for large reasoning models in image restoration and propose VLCoT-GRPO, a multi-modal RL framework that coordinates and optimizes both visual and textual reasoning for more coherent and faithful RealSR reconstruction.
%To address this gap, our work is the first to explore reinforcement learning for large reasoning models in the image restoration field.

\section{Method}
\subsection{Preliminary}
%\subsubsection{CoT}
%\subsubsection{GRPO}
Group Relative Policy Optimization (GRPO)~\cite{shao2024deepseekmath} is a reinforcement learning framework aimed at improving inference performance in large-scale models. In contrast to conventional approaches such as Proximal Policy Optimization (PPO)~\cite{schulman2017proximal}, which rely on an explicit value function for variance reduction, GRPO employs a baseline constructed from the mean reward of multiple rollouts conditioned on the same input. This design eliminates the dependency on auxiliary value estimation while maintaining stable and efficient policy updates. Given a question-answer pair $(q, a)$, a set of $G$ responses $\{o_i\}_{i=1}^{G}$ is drawn from the prior policy $\pi_{\theta_{\text{old}}}$. Subsequently, the policy model $\pi_{\theta_{\text{old}}}$ is optimized by maximizing the objective function:
\begin{equation}
\footnotesize
\label{grpo}
\begin{aligned}
&\mathcal{J}_{GRPO}(\theta) = \mathbb{E}\left[(q, a) \sim D,\ \{o_i\}_{i=1}^G \sim \pi_{\theta_{\text{old}}}(O \mid q)\right] \\
&\frac{1}{G} \sum_{i=1}^G \Bigg(
    \min \Bigg(
        \frac{\pi_\theta(o_i \mid q)}{\pi_{\theta_{\text{old}}}(o_i \mid q)} A_i, \operatorname{clip}\!\left(
        \frac{\pi_\theta(o_i \mid q)}{\pi_{\theta_{\text{old}}}(o_i \mid q)},\ 
        1-\varepsilon,\ 1+\varepsilon
    \right) A_i \Bigg)\\
& \quad  - \beta \, \mathbb{D}_{KL}\!\left(\pi_\theta \,\|\, \pi_{\text{ref}}\right)
\Bigg).
\end{aligned}
\end{equation}
GRPO adopts a clipped surrogate objective similar to that used in Proximal Policy Optimization (PPO), where $\varepsilon$ and $\beta$ are hyperparameters controlling the clipping range and regularization strength, respectively. In addition, GRPO explicitly incorporates a KL divergence penalty between the current policy $\pi_{\theta}$ and a reference policy $\pi_{\theta_{\text{ref}}}$ into the loss function:
\begin{equation}
\mathbb{D}_{K L}\left(\pi_\theta \| \pi_{r e f}\right)=\frac{\pi_{r e f}\left(o_i \mid q\right)}{\pi_\theta\left(o_i \mid q\right)}-\log \frac{\pi_{r e f}\left(o_i \mid q\right)}{\pi_\theta\left(o_i \mid q\right)}-1,
\end{equation}
where, $A_i$ denotes the relative advantage of the $i$-th response, which is obtained by feeding each sampled response into a reward model to compute its individual reward $r_i$. The set of rewards $\{r_1, r_2, \cdots, r_G\}$ is then normalized across the group to compute the advantage values:
\begin{equation}
A_i=\frac{r_i-\operatorname{mean}\left(\left\{r_1, r_2, \cdots, r_G\right\}\right)}{\operatorname{std}\left(\left\{r_1, r_2, \cdots, r_G\right\}\right)}.
\end{equation}

The reward functions are rule-based, evaluating the correctness of the final answer and output format without imposing constraints on the intermediate reasoning steps.

\begin{figure*}[h]
    \centering
    \includegraphics[width=0.99\textwidth]{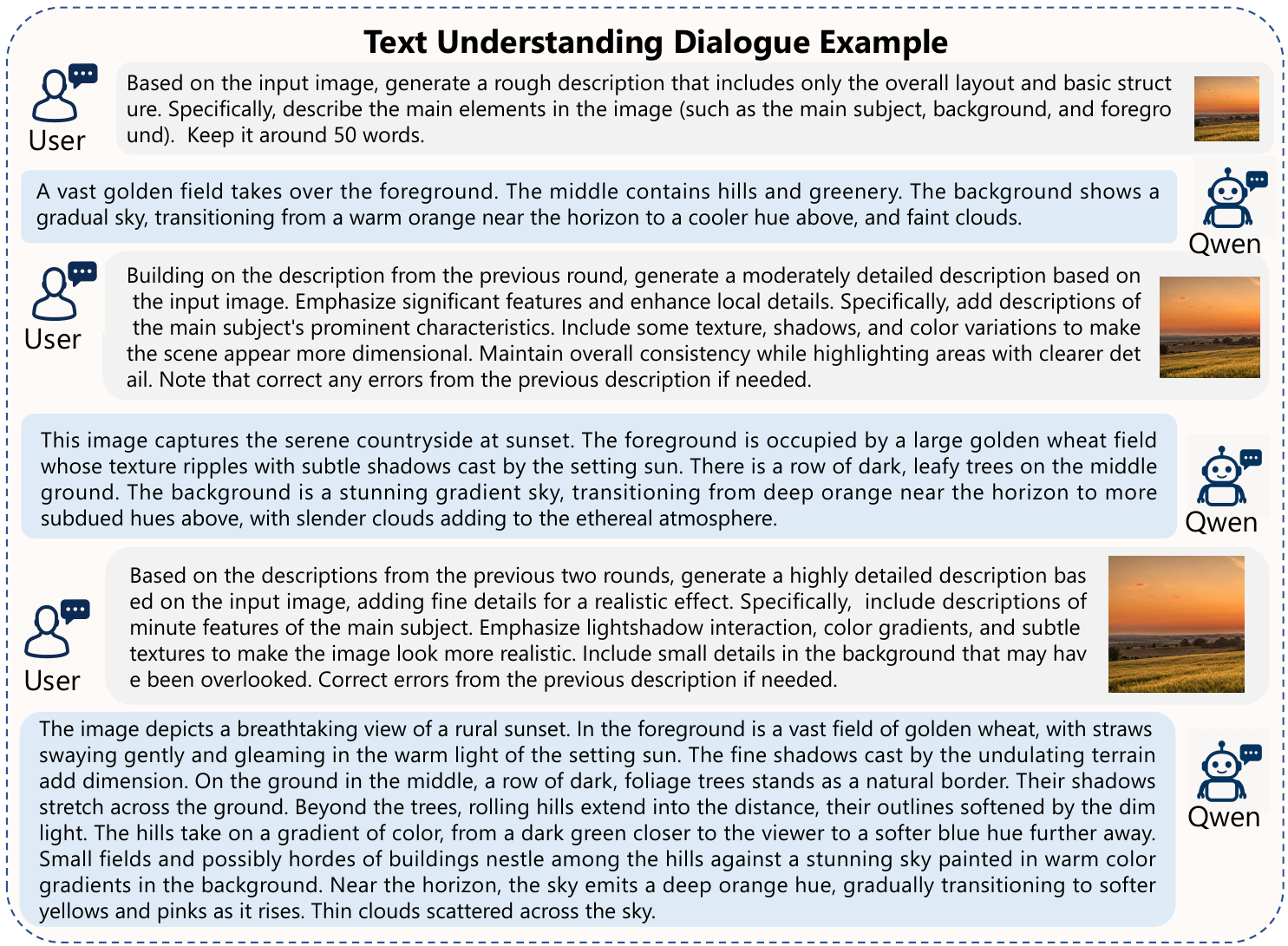} 
    \caption{Visual example of step-by-step generation of detailed image descriptions.}
    \label{fig:dia}
    \vspace{-0.5cm}
\end{figure*}

\subsection{VLCoT}
\textbf{VLCoT process design.}
Unlike traditional single-modal Chain-of-Thought design, human image restoration does not separate the processes of understanding and generation. Therefore, we propose Vision-Language Chain of Thought (VLCoT), a coarse-to-fine, multi-step reasoning process that integrates both textual and visual information. As illustrated in Fig.~\ref{fig:method}, given an LR image, we guide the model to generate VLCoT reasoning with the following prompt: \textit{`Perceive the degradation, understand the image content, and restore the high-quality image step by step (simulating the image restoration process in three steps from coarse to fine).'} Then VLCoT decomposes the image restoration process into a sequence of multimodal reasoning steps, including: (1) degradation perception step; (2) coarse restoration step (coarse understanding and image generation); (3) middle-detail restoration step  (middle-detail understanding and image generation); and (4) fine-detail restoration step (fine-detail understanding and image generation). This design mimics the human perception-to-reasoning pathway, starting from degradation awareness, followed by a coarse-to-fine process of understanding and image generation, enabling the model to progressively and robustly restore both semantic and structural information from severely degraded images.

\textbf{Cold-start Initialization for VLCoT.}
To ensure that the model has stable output style and basic image restoration capabilities from the start, we perform cold-start training on VLCoT before fine-tuning the entire model using reinforcement learning. Specifically, we first construct paired input-output and CoT datasets, which will be detailed in the following sections, and process the concatenated text-image token sequences through autoregressive learning. Given a multimodal sequence $X=\left(x_1, x_2, \ldots, x_T\right)$, where each $x_t$ may be a text token, image patch, or other modality embedding, the learning objective is to maximize the likelihood of the next element conditioned on previous observations:
$P_{\left(x \mid y \right)}(\theta)=-\sum_{t=1}^T \log P_\theta\left(x_t \mid x_{<t}, I\right)$
where I is the input and $x_{<t}=\left(x_1, \ldots, x_{t-1}\right)$ represents the preceding context.
The overall training optimization procedure minimizes the stepwise Cross-Entropy loss between the predictions and the targets:
\begin{equation}
\label{train}
\mathcal{L}=\operatorname{CrossEntropy}\left(g_t, P_\theta\left(x_t \mid x_{<t}, I\right)\right),
\end{equation}

For the overall framework, we utilize Lumina-mGPT~\cite{liu2024lumina} as our foundation, a strong MLLM model based on a pretrained decoder-only Transformer architecture.

\textbf{Degradation perception step.}
In the degradation perception step, we follow~\cite{wang2021real,wu2024seesr} by randomly selecting degradation (e.g., Gaussian or Poisson noise) and sampling all parameters (e.g., noise intensity $\mu_1 \in[a, b]$) from a uniform distribution to determine the degradation. However, these degradation ground truth (GT) values with specific parameters are difficult for MLLMs to comprehend. Therefore, we further transform them into user-friendly summary text representations.
Specifically, we follow~\cite{zhang2023crafting,chen2023image} to discretize the sampling distribution of the parameters for each degradation component (e.g., noise, blur, JPEG artifacts) by evenly dividing it into discrete intervals. These intervals are then discretely described and summarized to represent the degradation level. For example, we divide the distribution of noise level $\varphi_1$ into three uniform intervals, which are described as "light," "medium," and "heavy." Finally, the overall degradation representation combines the descriptions of all components, such as "[deblurring description, noise description,..., resizing description]".

\textbf{Restoration step.}
In the image generation stage of each restoration step, the ground-truth supervision for the coarse, middle-detail, and fine-detail images is derived from the HQ image downsampled by factors of 4 and 2, as well as the original resolution, respectively.
For the understanding text generation, we utilize Qwen2.5-VL~\cite{bai2025qwen2} to produce descriptions of the GT images at each level (coarse, middle-detail, and fine-detail restoration images) through a three-round dialogue process. The detailed prompts used in the dialogue rounds are provided in the Fig~\ref{fig:dia}. Each round is guided by a specific
designed prompts to elicit increasingly fine-grained textual understanding of the visual content.

\textbf{Tokenization.}
For RealSR tasks, high compression ratios and limited codebook capacity result in significant information loss during the extraction of discrete image representations. To mitigate this issue, we adopt the VQGAN tokenizer from LlamaGen~\cite{sun2024autoregressive}, which offers a lower downsampling factor compared to the original Lumina-mGPT (e.g., $8 \times$ \emph{vs.} $16 \times$) and a larger visual vocabulary size (e.g., 16,536 \emph{vs.} 8,192). This modification substantially reduces information loss during the encoding process, thereby improving the quality of reconstructed images in the Real-ISR task. For text tokenization, we retain the BPE tokenizer used in Lumina-mGPT, which originally has a vocabulary size of 65,536. To accommodate the expanded image tokenizer, the BPE tokenizer's vocabulary size is increased to 73,728, enabling it to handle a larger set of discrete image tokens.

\begin{figure*}[t]
    \centering
    \includegraphics[width=0.98\textwidth]{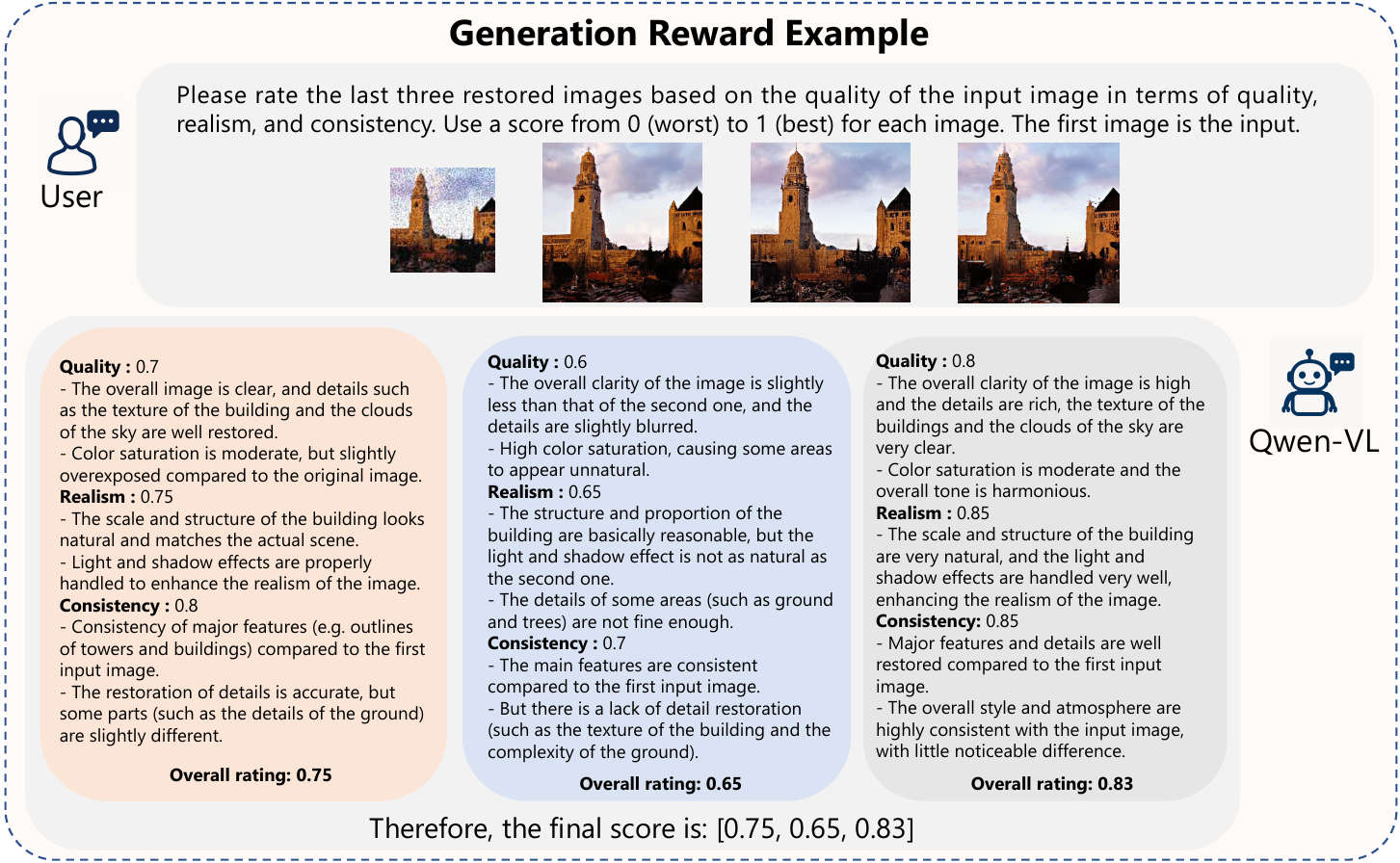} 
    \caption{Illustrative example of a vision expert model assigning scores to a set of images.}
    \label{fig:score}
    \vspace{-0.5cm}
\end{figure*}

\subsection{VLCoT-GRPO}
Reinforcement learning has been proven effective in enhancing reasoning capability in large language models, as demonstrated by the GRPO~\cite{guo2025deepseek}. To enable cross-modal collaborative reasoning in RealSR, we propose VLCoT-GRPO. This reinforcement learning framework optimizes visual and textual reasoning processes within the VLCoT. Compared with pure text-based reasoning tasks, multi-modal CoT reinforcement presents unique challenges. Due to the inherent limitation of current training, most MLLMs cannot natively perform alternating image-text generation. The model must dynamically determine when to reason for semantic understanding or image restoration. Traditional supervised fine-tuning rely on labeled prompts specifying which modality to generate, whereas VLCoT-GRPO introduces implicit task coordination via reinforcement learning.

For a given query $q$, VLCoT-GRPO performs iterative vision--language reasoning over $T$ steps. 
At each step $t$, the model alternates between textual reasoning and visual reconstruction, producing an interleaved output sequence $o_i = \{ l_i, v_i \}$, 
where $l_i$ denotes the $i^{th}$ languag token and $v_i$ denotes the corresponding vision token. 
Each reasoning step refines both semantic understanding and visual reconstruction conditioned on all previous steps.

Following the GRPO framework, the policy ratio between the current and old policies at step $t$ is defined as:
\begin{equation}
\begin{cases}
\dfrac{\pi_\theta(l_{l,j}\mid q, o_{<t}, l_{t,<j})}
{\pi_{\theta_{\text{old}}}(l_{l,j}\mid q, o_{<t}, l_{t,<j})}, 
& 0 \le j \le |l_t|, \\[6pt]
\dfrac{\pi_\theta(v_{l,j-|l_t|}\mid q, o_{<t}, l_t, v_{t,<j-|l_t|})}
{\pi_{\theta_{\text{old}}}(v_{l,j-|l_t|}\mid q, o_{<t}, l_t, v_{t,<j-|l_t|})}, 
& |l_t| < j \le |l_t| + M_t,
\end{cases}
\label{eq:ratio_iter}
\end{equation}
where $M_t$ represents the number of visual tokens generated in step $t$.

\subsection{Reward Functions}
To stimulate the model's self-exploration and enhance its reasoning ability, we propose VLCoT-GRPO. As shown in Fig.~\ref {fig:method}, VLCoT-GRPO is equipped with a set of carefully designed reward functions aimed at improving the stability and effectiveness of the multi-step, multi-modal reasoning process.
Specifically, our reward $R$ consists of four parts, including the format reward, degradation reward, understanding reward and generation reward:
\begin{equation}
R=R_{form}+R_{deg}+R_{und}+R_{gen}
\end{equation}
These rewards are designed to address challenges such as inaccurate degradation perception, misunderstandings, hallucinated content, and poor restoration quality.

\textbf{Format Reward.}
To enforce the structural integrity of the multi-step reasoning process in VLCoT, we introduce a format reward that guides the model to generate outputs with predefined tags (e.g., \textit{\textless degradation\textgreater...\textless/degradation\textgreater...}). We use rule-based checks to ensure each step is properly wrapped in corresponding tags with correct order and completeness. A reward of $1$ is given for well-formed outputs; otherwise, $0$.

\textbf{Degradation Reward.}
The Degradation Reward is used to supervise the model's output and ensure the correctness of the degradation estimation. Specifically, we compare the output with the degradation ground truth obtained from the cold-start phase degradation estimation pipeline. This comparison checks whether each discrete degradation components are correctly identified and assigns scores based on the accuracy of the degradation output. The degradation reward $R_{deg}$ can be formalized as $R_{deg} = \frac{D_{correct}}{D_{total}}$, where $D_{correct}$ and $D_{total}$ represent the number of correctly estimated degradations and the total number of discrete degradation components, respectively.

\textbf{Understand Reward.}
In VLCoT, the coarse understanding plays a key role in identifying the main objects and layout of the image. Once the main objects are accurately recognized at this step, the subsequent middle- and fine-detail restoration steps can focus on detail enhancement and quality refinement based on the established coarse understanding and image. To encourage the model to build a clear and comprehensive global understanding early on, we design the Understand Reward $R_{und}$, which measures how well the coarse understanding text covers the main objects in the input image. 
Specifically, we employ Grounded-SAM~\cite{ren2024grounded}, which supports open-vocabulary detection and segmentation, to detect main objects in the input LR images. This process extracts a set of key objects present in the scene.
Subsequently, we check whether these main objects are mentioned in the coarse understanding text. If an object is explicitly referenced in the text, it is considered a hit. The Understanding Reward is defined as follows:$R_{und} = \frac{N_{cover}}{N_{total}}$, where $N_{cover}$ denotes the number of main objects correctly mentioned in the coarse understanding text, and $N_{total}$ represents the total number of main objects.

\textbf{Generation Reward.}
In contrast to the rule-based reward design in GRPO, the evaluation of image restoration quality in the RealSR task typically involves multiple dimensions, including perceptual visual quality, alignment, and multi-scale structural integrity. A single predefined rule is insufficient to capture these diverse aspects.
To comprehensively assess the image restoration quality, we leverage a powerful prior visual expert model (e.g., Qwen2.5-VL~\cite{bai2025qwen2}) as a visual expert to conduct a multi-faceted evaluation of the output images, considering factors such as quality, realism, and consistency. Fig.~\ref{fig:score} illustrates the scoring process of generated images. Alongside the individual scores, the expert provides interpretability through a rationale for each assessment. Finally, the scores across all dimensions are aggregated via a weighted sum to produce an overall image quality score. It is worth mentioning that experts judge the quality of a group of multiple restored images simultaneously. This is because we encourage the model to learn relative preferences instead of memorizing fixed quality values. This comparative learning paradigm helps RealSR-R1 develop its own internal "value system" for generation quality, rather than simply overfitting to the biases of the reward model.

\section{Experiments}
\label{exper}

\subsection{Experimental Setting}
\textbf{Training and Testing Datasets.} For cold-start stage, we adopt the full LSDIR dataset~\cite{li2023lsdir} and the first 10k face images from the FFHQ dataset~\cite{karras2019style} as training dataset. Then, based on Equ.~\ref {grpo}, we select 500 high-quality images and text pairs for training our RealSR-R1. To generate the low-quality (LR) images, we employ the degradation process of Real-ESRGAN~\cite{wang2021real} and use the same degradation settings as in SeeSR~\cite{wu2024seesr} to ensure a fair comparison. For testing datasets, following previous works~\cite{wang2024exploiting,dong2024tsd,qu2025visual,wu2024one}, we use three real-world datasets and one synthetic dataset for evaluation. For the real-world datasets, we test on the Dreal~\cite{wei2020component} and real~\cite{cai2019toward} testset provided by Osediff~\cite{wu2024one}, as well as the RealLR250~\cite{ai2024dreamclear} dataset. Following PURE~\cite{wei2025perceive}, we use the OST-Val~\cite{wang2018recovering} dataset, processed with the same degradation process in SeeSR~\cite{wu2024seesr}, as the synthetic test set. %The LR-HQ pairs for all testing datasets have sizes of $128 \times 128$ and $512 \times 512$, respectively.

\input{table/main_compare}

\textbf{Evaluation metrics.}
To comprehensively evaluate the performance of different methods, we combine full-reference and no-reference metrics. PSNR and SSIM~\cite{wang2004image} (calculated on the Y channel in the YCbCr color space) are full-reference fidelity measures, while LPIPS~\cite{zhang2018unreasonable} and DISTS~\cite{ding2020image} are full-reference perceptual quality metrics. FID~\cite{heusel2017gans} assesses the distributional distance between the original and reconstructed images. For blind quality assessment of no-reference images, we combine NIQE~\cite{zhang2015feature}, MANIQA~\cite{yang2022maniqa}, MUSIQ~\cite{ke2021musiq}, CLIPIQA~\cite{wang2023exploring}, and TOPIQ~\cite{chen2024topiq} to evaluate the naturalness and structural integrity of the images.

\textbf{Implementation Details.}
In the cold-start phase, we use the AdamW optimizer with an initial learning rate of $2 \times 10^{-5}$, and employ a cosine annealing strategy for learning rate decay. The model is trained for 2 epochs on the entire dataset with batchsize=16, including a warmup phase spanning the first 0.01 epochs to stabilize early-step optimization. In the VLCoT-GRPO phase, the AdamW optimizer is again used with an initial learning rate of $1 \times 10^{-5}$, a batch size of 8, and a constraint on the maximum generated token length of $6k$. For the same LR image, RealSR-R1 samples three responses and provides corresponding rewards. The model was trained for 2 epochs.

\subsection{Comparison with State-of-the-Arts}
\textbf{Quantitative Comparisons.}
The quantitative comparison is shown in Tab.~\ref{tab:methods}. On synthetic datasets, RealSR-R1 achieves the best or second-best results across seven metrics, reflecting the capability to generate high-quality and realistic images. On real-world benchmarks, our method performs strongly across most no-reference metrics (e.g., NIQE~\cite{zhang2015feature}, MANIQA~\cite{yang2022maniqa}, MUSIQ~\cite{ke2021musiq}, CLIPIQA~\cite{wang2023exploring}, and TOPIQ~\cite{chen2024topiq}), demonstrating its strong ability to generate more realistic images. It is worth noting that although VARSR~\cite{qu2025visual} achieves competitive performance in image generation, its training relies on proprietary data. In contrast, RealSR-R1 surpasses VARSR in generation metrics while being trained solely on publicly available datasets. However, PSNR and SSIM still have a gap with diffusion-based methods. This is due to the trade-off between realism and fidelity, as RealSR-R1 generates more textures and details, which may reduce fidelity metrics. Similarly, recent state-of-the-art methods such as PURE~\cite{wei2025perceive} and VARSR also struggle to simultaneously achieve the best PSNR and SSIM while maintaining strong perceptual quality. In contrast, RealSR-R1 provides greater controllability. By adjusting the orientation of the generation reward within VLCoT-GRPO, the model can be guided toward fidelity-oriented reconstruction. A more detailed analysis of this trade-off and the corresponding ablation experiments are presented in Tab.~\ref{tab:G_reward} in Sec.~\ref{aba}.

%Recent studies~\cite{you2024depicting, yu2024scaling, ai2024dreamclear} have shown that the restored images exhibit higher quality in terms of human perception, but perform poorly on certain reference-based metrics (e.g., PSNR and SSIM). Therefore, it is necessary to reconsider the reference values of existing metrics and propose more effective methods to evaluate modern image restoration methods. 

\begin{figure*}[t]
    \centering
    \includegraphics[width=0.99\textwidth]{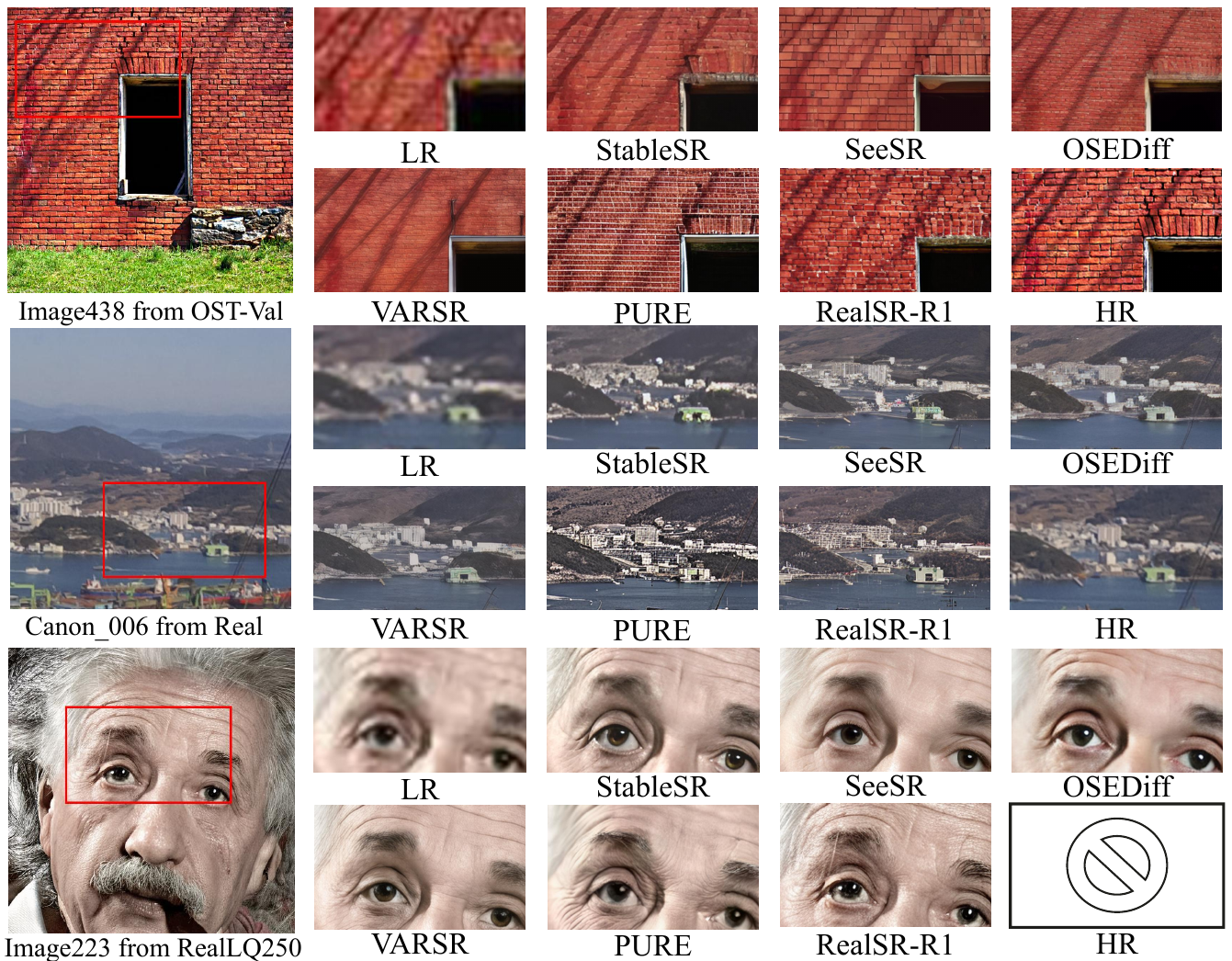} 
    \caption{Qualitative comparisons with different SOTA methods.}
    \label{fig:visul}
    %\vspace{-0.7cm}
\end{figure*}

\textbf{Qualitative Comparisons.}
In Fig.~\ref{fig:visul}, we present a visual comparison of different datasets. As shown in the first case, even when faced with severely degraded input, RealSR-R1 generates realistic wall texture details, while other methods produce overly smooth or incorrect textures. In the second example, SeeSR and OSEDiff produce blurred structures in the background buildings, while PURE introduces excessive pseudo-textures. In contrast, our method most accurately restores the image details. On the RealLR250 dataset, RealSR-R1 also demonstrates an advantage in visual quality, generating more realistic face details while maintaining better overall image fidelity.

\begin{figure}[h]
  \centering
   \includegraphics[width=1.\linewidth]{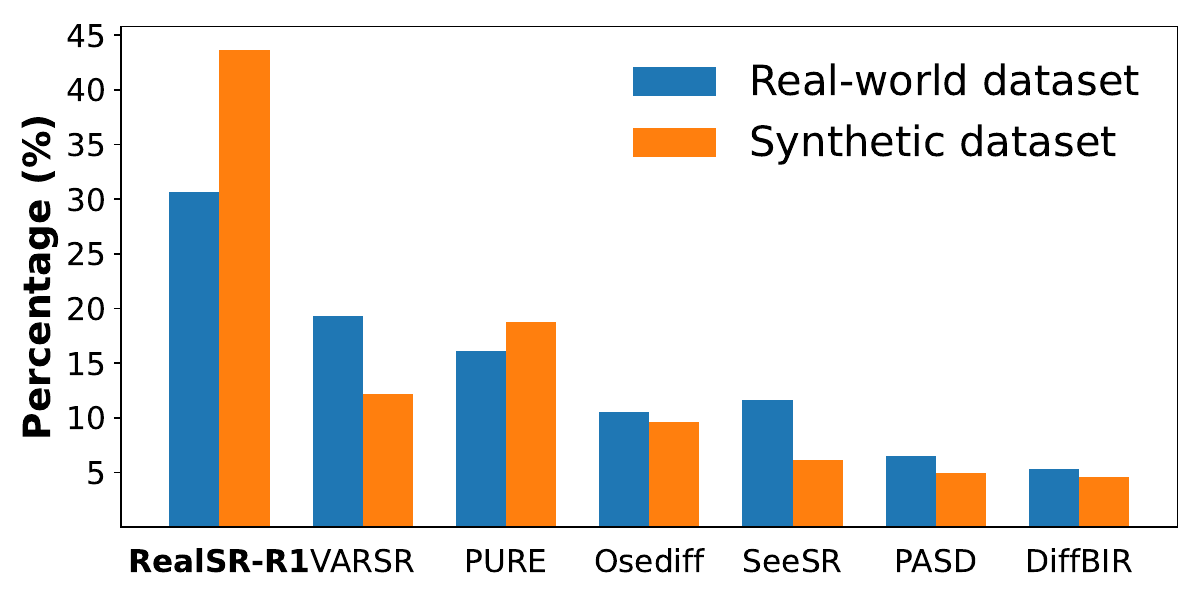}
   \vspace{-1em}
   \caption{User study results on real-world dataset and synthetic dataset.}
   \vspace{-1em}
   \label{fig:user}
\end{figure}

\textbf{User Study.}
To further demonstrate that the images generated by our method align more closely with human perception, we conducted a user study with 20 participants, evaluating both synthetic and real-world data. Specifically, we randomly selected 50 real-world LR images from RealSR, DrealSR, and RealLR250, and chose 25 degraded images from the synthetic dataset OST-Val. We then compared the enhancement results of RealSR-R1 with six methods (PURE, SeeSR, VARSR, PASD, DiffBIR, and OSEDiff). Participants were asked to vote on the best restoration results based on visual quality, naturalness, and accuracy. Importantly, to ensure reflects real-world deployment conditions, we did not provide the ground-truth for participants. This no-reference setup aligns with practical scenarios where end users typically have access only to the degraded input and model outputs, not the ground-truth. We believe this design better captures human perceptual preference and the practical utility of each method. As shown in Fig.~\ref{fig:user}, our RealSR-R1 achieved better preferences across different datasets than other methods, with a selection rate of 30.7\% and 43.6\%, respectively. This result demonstrates its advantage in generating high-quality images.

\begin{figure*}[t]
    \centering
    \includegraphics[width=0.9\textwidth]{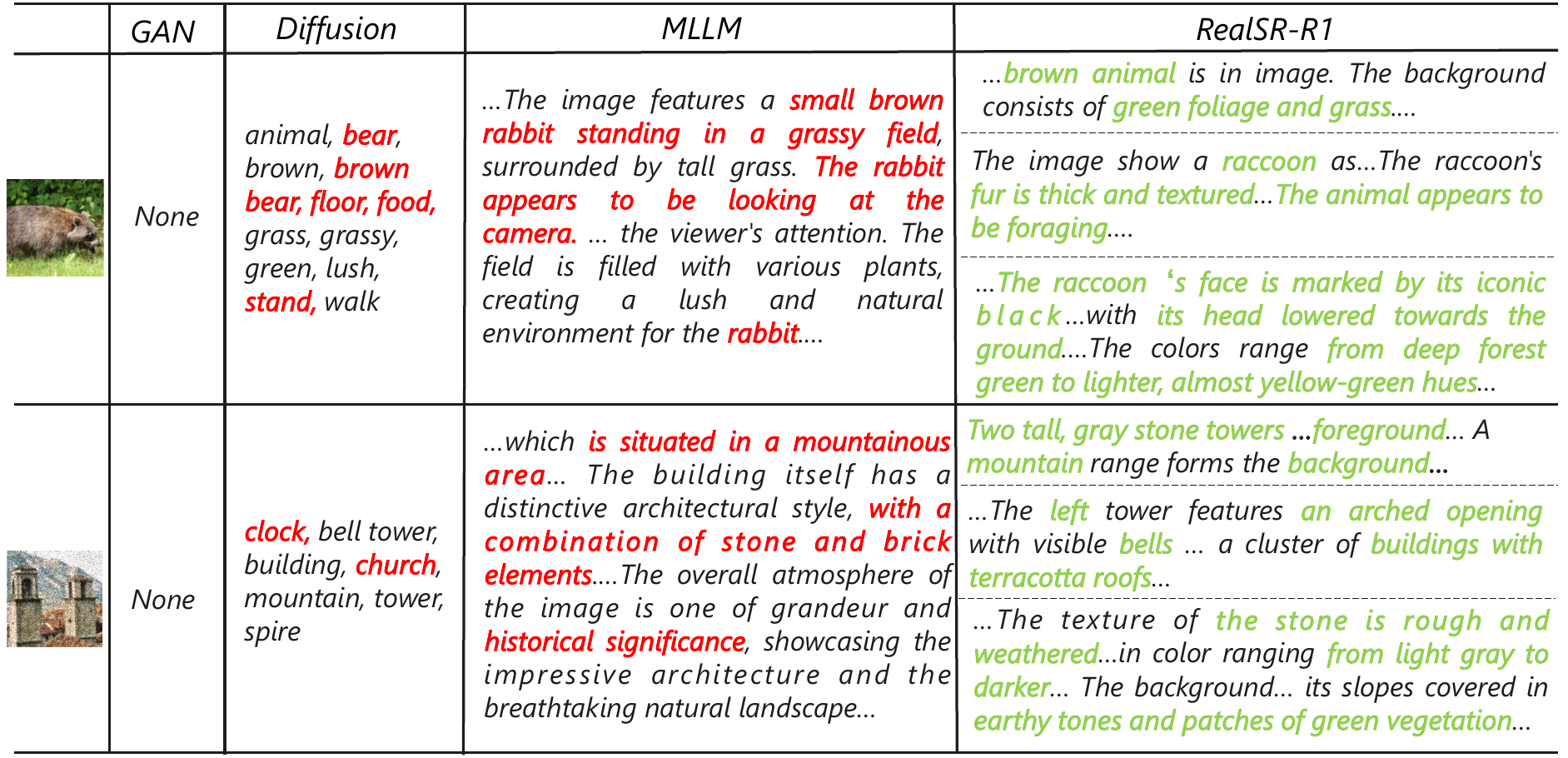} 
    \vspace{-2mm}
    \caption{Comparison of the understanding ability of different methods on real SR tasks.}
    \label{fig:understand}
    \vspace{-0.1cm}
\end{figure*}

\textbf{Understanding Capabilities of Different Models}
We further compared the understanding capabilities of GAN-based, Diffusion-based, and MLLM-based methods for real image super-resolution. As shown in Fig.~\ref{fig:understand}, GAN-based methods lack an understanding of image content, while diffusion-based methods utilize external language models (e.g., DAPE~\cite{wu2024seesr}) to extract simple image descriptions as prompts. MLLMs possess more refined understanding capabilities, but directly applying this fine-grained understanding to severely degraded images leads to incorrect interpretations of details. For example, diffusion and MLLM mistakenly interpret the image as a  bear and a rabbit, respectively. In contrast, only RealSR-R1 correctly recognizes the image as a raccoon and generates the most accurate and detailed image description, which is crucial for restoring image details.

\textbf{Model Complexity Comparison with PURE} 
We have conducted a complexity and efficiency comparison between our RealSR-R1 model and PURE~\cite{wei2025perceive}, a representative autoregressive multimodal generation method. All evaluations were performed using $512 \times 512$ input images on an NVIDIA A800 GPU under consistent conditions.

\input{table/rebuttal_time}

As shown in Tab.~\ref{tab:re_time}, RealSR-R1 has the same number of parameters as PURE. The inference time of our model is slightly longer (+56s), primarily due to the inclusion of coarse and middle-detail stages. However, this design substantially enhances the model's visual understanding and generation capabilities with the help of R1, leading to notable improvements across all ten evaluation metrics, which is very difficult, so the slight sacrifice in speed is acceptable. 
It is also worth highlighting that, to the best of our knowledge, our work is the first to successfully adapt an R1-style architecture to a low-level vision task and empirically demonstrate its effectiveness. We believe this provides a valuable foundation for future exploration of autoregressive reasoning in image restoration.

\textbf{Extension to Image Deraining Task}
To further evaluate the generalizability of RealSR-R1 to other image restoration tasks, we conducted an experiment on image draining. Specifically, we generated a deraining dataset by adding synthetic raindrops to clean training images. We then fine-tuned RealSR-R1 for only 10K iterations (including cold-start and VLCoT-GRPO) on this task. Despite the limited training, the model still shows remarkable performance, as shown in Tab.~\ref{tab:re_rain}.

\input{table/rebuttal_rain}

These results demonstrate that VLCoT’s step-wise visual reasoning and the VLCoT-GRPO training strategy can benefit other image restoration tasks.

\subsection{Ablation Study}
\label{aba}
\input{table/aba_mian}
\textbf{Ablation of the key components of RealSR-R1.}
To validate the effectiveness of RealSR-R1, we systematically conducted ablation experiments on OST-Val. We progressively removed core components from RealSR-R1. The results, as shown in Tab~\ref{tab:CoT}, removing the VLCoT GRPO leads to significant reductions in all metrics, demonstrating the substantial potential of R1 in real-world super-resolution. Secondly, the degradation and understanding output affect different metrics: removing the degradation output and removing the understanding output significantly reduce the reference-based metrics SSIM, LPIPS, and the no-reference metric MANIQA, respectively. Furthermore, removing multi-step outputs and reducing the length of CoT performs poorly on most metrics. These observations validate the effectiveness of each core component of the proposed method.

\textbf{Ablation of VLCoT-GRPO and reward functions.}
We further performed a step-by-step removal of the proposed reward functions to ablate their effectiveness. In Tab.~\ref{tab:reward}, we find that different rewards affect different metrics. For example, the generation reward significantly improves CLIPIQA and TOPIQ, while the degradation reward tends to improve SSIM and LPIPS. Additionally, removing the understanding reward impacts image restoration quality across all metrics, indicating that semantic understanding plays a crucial role in driving the generation of real-world details.

\textbf{The ablation of understanding reward.}
We conduct ablation studies on tag extraction strategies within the understanding reward to evaluate the impact of different tag extraction methods on model performance. As shown in Tab~\ref{tab:U_reward}, RAM~\cite{zhang2024recognize} performs poorly on both perceptual and generative metrics, which may be attributed to the fact that the extracted tags often contain incorrect or misleading information. In contrast, the tags extracted by Ground-SAM achieve the best performance across all four metrics. Therefore, we adopt Ground-SAM as the tag extraction method for the understanding reward in RealSR-R1.
\input{table/aba_reward_U}

\textbf{The ablation of generation reward.}
We further conduct ablation studies on the visual expert used in the generation reward to verify the controllability of RealSR-R1. As shown in Tab.~\ref{tab:G_reward}, using perceptual metrics as the reward leads the model to prioritize reconstruction fidelity—achieving the highest PSNR and SSIM. In contrast, employing generative quality metrics drives the model toward visually appealing and realistic details, obtaining the best MANIQA and TOPIQ scores. These results demonstrate that RealSR-R1 possesses controllable generation bias. By adjusting the reward orientation within the VLCoT-GRPO framework, the model can flexibly balance between reconstruction fidelity and perceptual realism, thereby achieving restoration outcomes that better align with different application requirements.

Notably, when using Qwen2.5-VL as the visual expert, the model achieves a balanced performance across all indicators—retaining relatively high PSNR/SSIM while improving perceptual metrics (e.g., CLIPIQA 0.7127, MANIQA 0.6858)—representing the most effective trade-off.
Moreover, unlike hand-crafted metrics that evaluate each aspect independently, Qwen2.5-VL jointly assesses perceptual and generative quality for the entire output set, providing more holistic and context-aware feedback for reinforcement learning optimization.
\input{table/reward_G}

\textbf{The ablation of different stage design.}
To further investigate the impact of our three-stage design, we conducted a comparison between three-step and two-step variants (removing the middle-detail stage). Results are shown in Tab.~\ref{tab:re_step}. 

\input{table/rebuttal_step}

These results demonstrate that the three-step reasoning process consistently provides better performance across all metrics. We hypothesize that this is because the two-step pipeline introduces a larger semantic gap between stages, making it harder for the model to progressively refine the image in a controllable manner.

\begin{figure}[h]
    \centering
    \includegraphics[width=0.95\columnwidth]{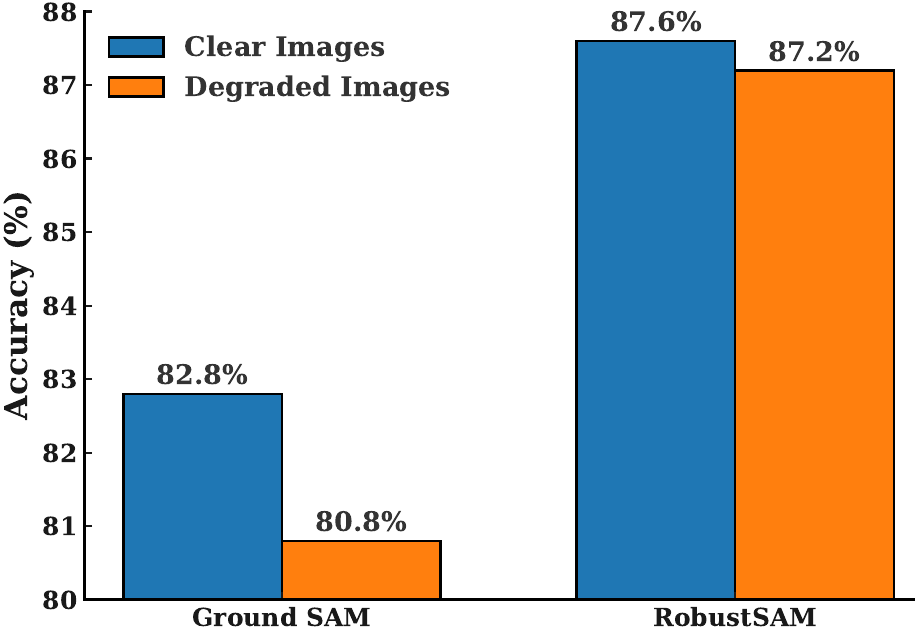} 
    % \vspace{-2mm}
    \caption{Comparison of detection accuracy of clean and degraded images using Grounded-SAM and Robust SAM.}
    \label{fig:sam}
    % \vspace{-0.1cm}
\end{figure}

\textbf{Performance of Grounded-SAM on low-quality images.}
To evaluate whether the performance of Grounded-SAM itself will degrade on low-quality images, we randomly selected 50 images from the RealSR and OST-Val test datasets and manually annotated their main objects. We then conducted a quantitative analysis of Grounded-SAM’s detection accuracy on the test dataset RealLQ250. As shown in the Fig.~\ref{fig:sam}, while Grounded-SAM does exhibit a drop in accuracy under degraded conditions, it still maintains a detection accuracy exceeding 80\%. To address this limitation further, we evaluated the performance of RobustSAM~\cite{chen2024robustsam}—a recent variant designed to enhance SAM's robustness on degraded images. Preliminary results show that RobustSAM achieves a +6.4\% improvement in accuracy over Grounded-SAM. We are actively exploring the integration of more resilient detectors to mitigate object recognition failures during training, and in the future will add confidence thresholds in reward formulation to minimize the risk of false penalties.

\section{Conclusion and Limitation}
In this paper, we propose a novel reinforcement Learning method for the real-world image super-resolution (RealSR) task, named RealSR-R1. The proposed Visual-Language Chain of Thought (VLCoT) framework effectively integrates vision and language, simulating human-like reasoning in image restoration, progressively refining low-resolution images into high-quality outputs. Furthermore, we propose VLCoT-GRPO and design reward functions to guide the model to ensure accurate degradation estimation, robust understanding, and precise generation of high-quality images. Extensive experiments and user studies demonstrate that RealSR-R1 outperforms existing state-of-the-art methods, achieving higher perceptual quality while maintaining alignment with human perception.

While our results are promising, we acknowledge that RealSR-R1 has some limitations. Our model relies on synthetic degradation ground truth (GT) as supervision for the degradation reward during training; however, GT for real-world images is often difficult to obtain. In future work, we will further investigate how to eliminate the need for GT images during training.

\section*{Acknowledgment}
This work is supported by the National Natural Science Foundation of China (NO. 62572193, 62102151), Shanghai Science and Technology Commission (22511104600), the Open Research Fund of the Key Laboratory of Advanced Theory and Application in Statistics and Data Science, Ministry of Education, and the Fundamental Research Funds for the Central Universities.
% \ifCLASSOPTIONcaptionsoff
%   \newpage
% \fi



%\bibliographystyle{IEEEtran}
\bibliographystyle{IEEEtran}
\bibliography{ref}


% \begin{IEEEbiography}{Shaohui Lin}
% \end{IEEEbiography}

% \begin{IEEEbiography}{Rongrong Ji}
% \end{IEEEbiography}

% % insert where needed to balance the two columns on the last page with
% %\newpage

% \begin{IEEEbiography}{Chao Chen}
% \end{IEEEbiography}

% \begin{IEEEbiography}{Feiyue Huang}
% \end{IEEEbiography}

% \begin{IEEEbiography}{Xuelong Li}
% \end{IEEEbiography}


%\vfill


\begin{IEEEbiography}[{\includegraphics[width=1in,height=1.25in,clip,keepaspectratio]{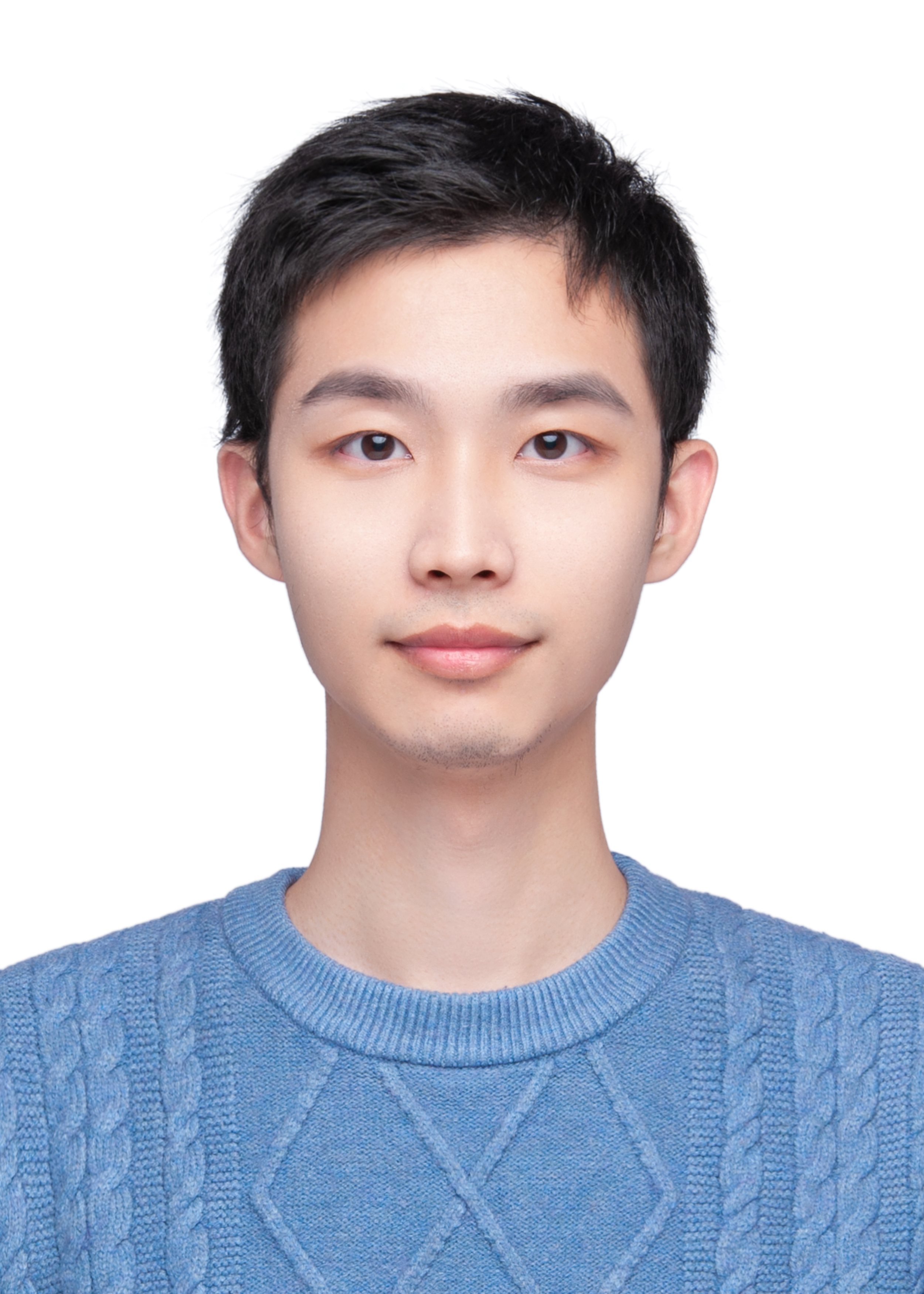}}]{Junbo Qiao}
received the B.S. degree from Zhengzhou University. He is currently pursuing the Ph.D. degree with the School of Computer Science and Technology, East China Normal University, Shanghai, China. His research interests include image restoration, multi-modal large language models, and AIGC.
\end{IEEEbiography}

\vspace{-12mm}

\begin{IEEEbiography}[{\includegraphics[width=1in,height=1.25in,clip,keepaspectratio]{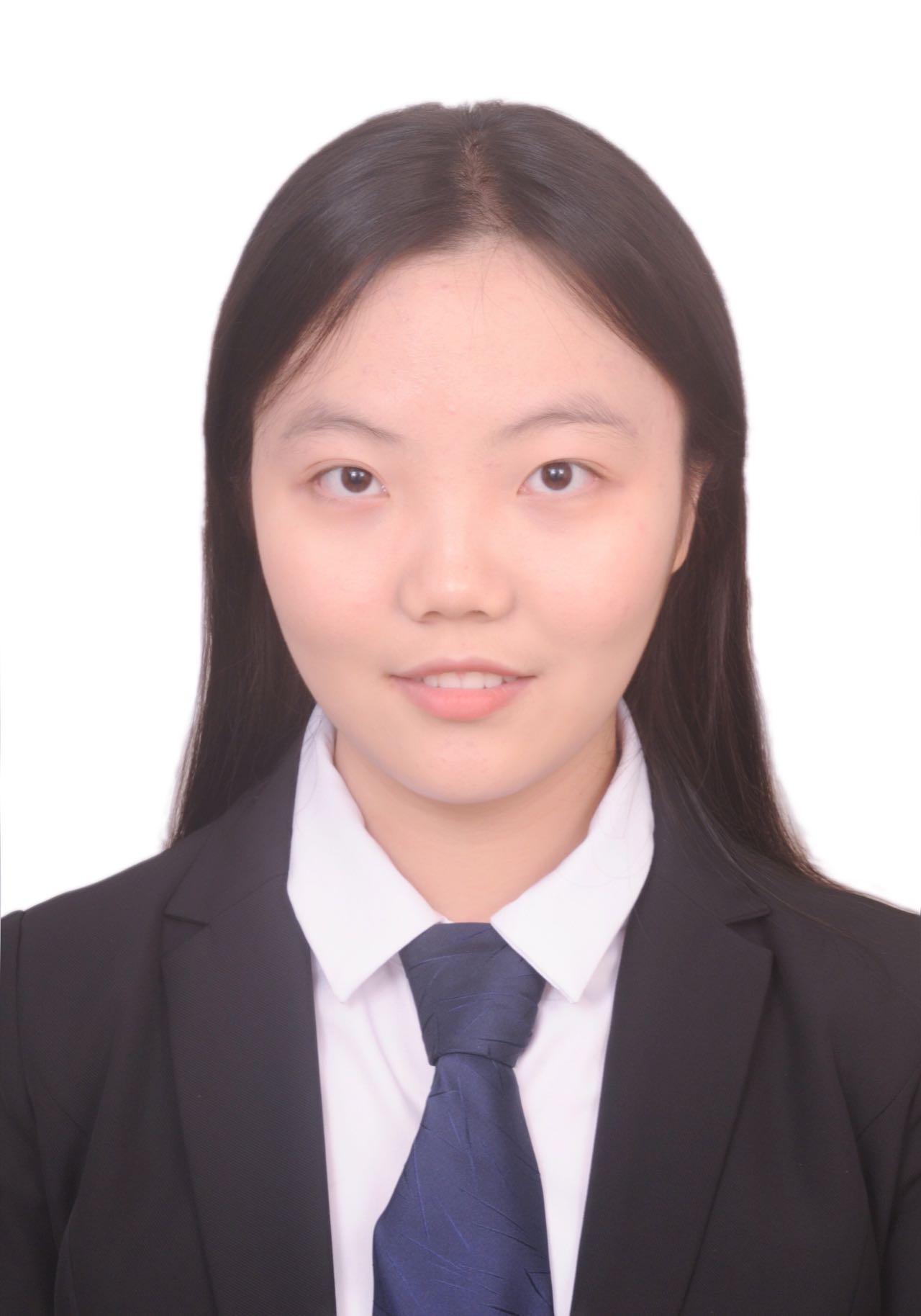}}]{Miaomiao Cai}
received the B.Sc. degree from Nanjing Agricultural University, Nanjing, China, in 2020. She is currently pursuing the Ph.D. degree in Information and Communication Engineering at the University of Science and Technology of China (USTC), Hefei, China. Her research has been published in prestigious journals and conferences, including the \textit{IEEE Journal of Biomedical and Health Informatics (JBHI)} and the \textit{International Conference on Medical Image Computing and Computer-Assisted Intervention (MICCAI)}. Her research interests include AI-generated content (AIGC), super-resolution, and image segmentation.
\end{IEEEbiography}

\vspace{-12mm}

\begin{IEEEbiography}[{\includegraphics[width=1in,height=1.25in,clip,keepaspectratio]{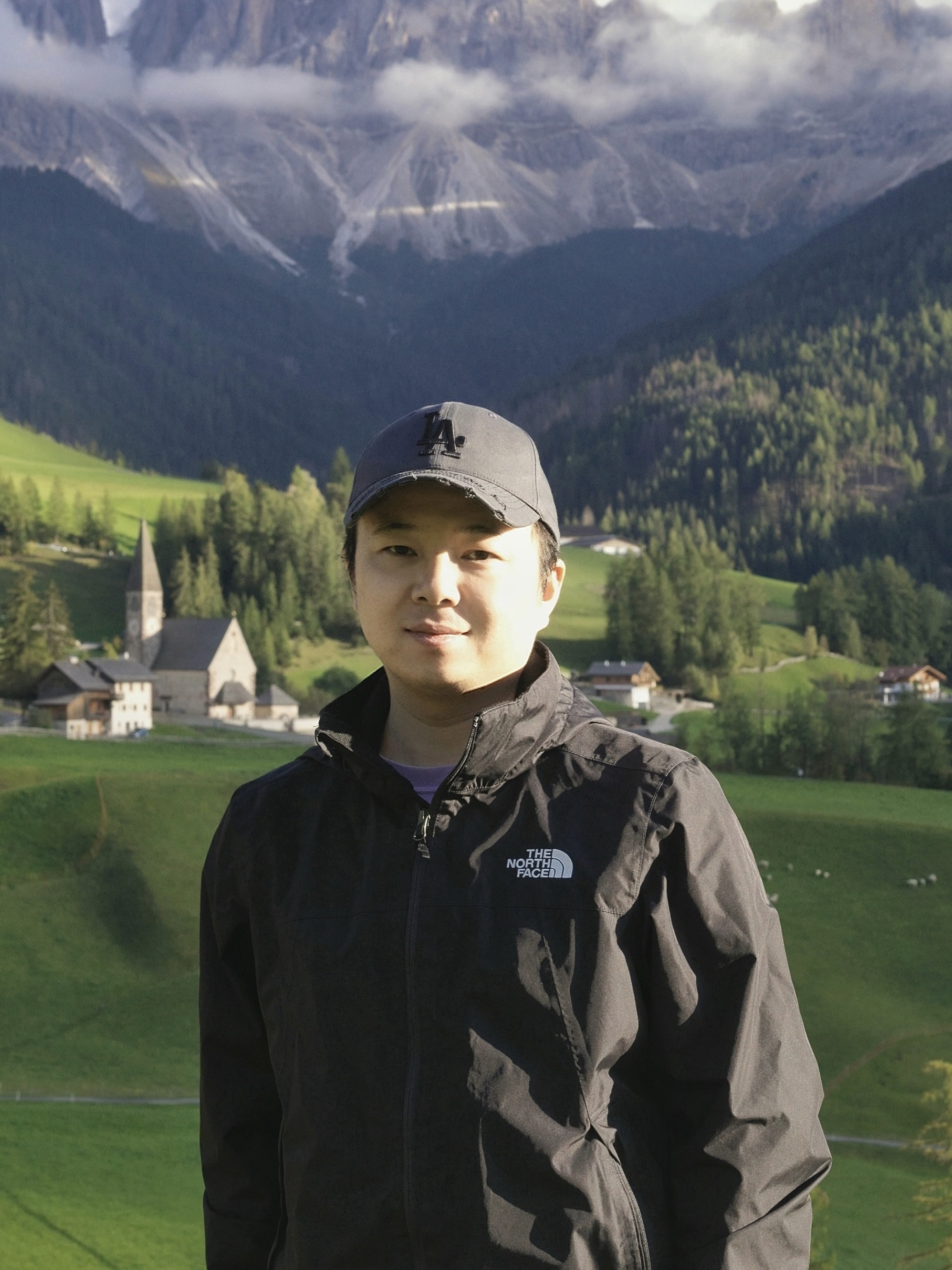}}]{Wei Li}
received his B.Sc. degree from Beijing Jiaotong University, Beijing, China, and M.Sc. degree from Tsinghua University, Beijing, China, in 2019. Presently, he is a Researcher at Huawei Noah’s Ark Lab. He has authored more than 10 scientific papers published in top-tier venues such as IEEE TIP, CVPR, ECCV, NeurIPS, ICLR, and AAAI. He also serves on the Program Committee for AAAI and as a reviewer for several prestigious journals and conferences, including IEEE TPAMI, TIP, NeurIPS, CVPR, and ICLR. His research interests primarily focus on machine learning and computer vision.
\end{IEEEbiography}

\vspace{-12mm}

\begin{IEEEbiography}[{\includegraphics[width=1in,height=1.25in,clip,keepaspectratio]{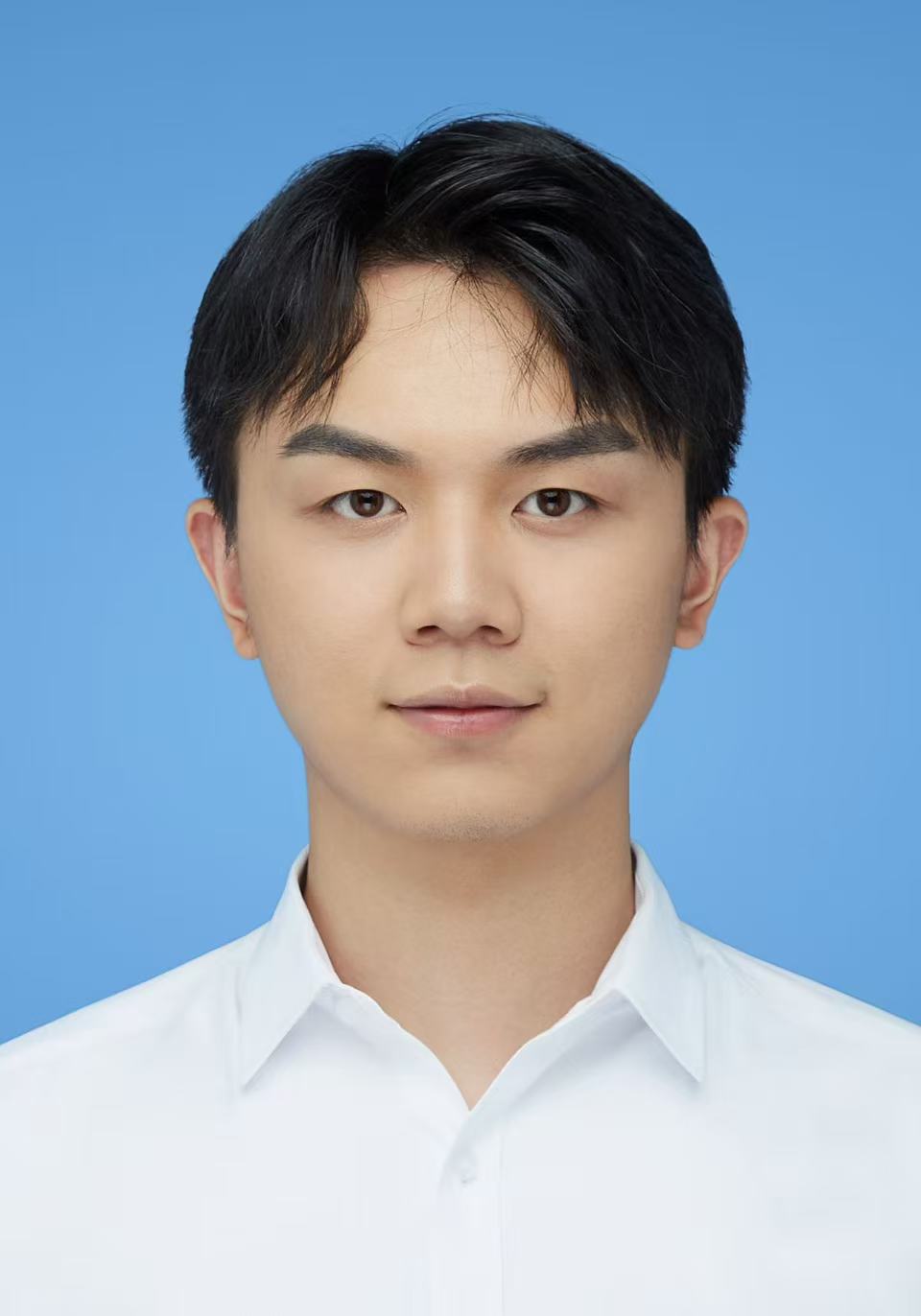}}]{Xudong Huang}
received his B.Sc. degree from Beijing Information Science and Technology University, Beijing, China, and M.Sc. degree from Beijing Jiaotong University, Beijing, China, in 2022. Currently, he is a Researcher at Huawei Noah’s Ark Lab. He has published papers in academic venues such as CVPR, NeurIPS, and AAAI. He also serves as a reviewer for several conferences, including CVPR, NeurIPS, and ICLR. His research interests include AIGC and reinforcement learning.
\end{IEEEbiography}

\vspace{-12mm}

\begin{IEEEbiography}[{\includegraphics[width=1in,height=1.25in,clip,keepaspectratio]{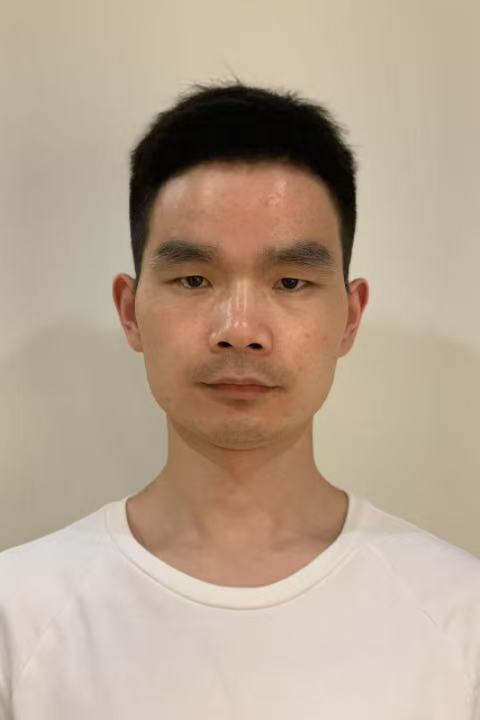}}]{Jie Hu}
received the B.Sc. degree from AnHui University, Hefei, China, in 2013, and the M.Sc. degrees from University of Science and Technology of China, Hefei, in 2016, respectively. He is currently a Researcher with HUAWEI Company. His research has been published in prestigious journals and academic venues, such as Transactions on Image Processing (TIP) and European Conference on Computer Vision (ECCV). His research interests include low-level vision and AIGC.
\end{IEEEbiography}

\vspace{-12mm}

\begin{IEEEbiography}[{\includegraphics[width=1in,height=1.25in,clip,keepaspectratio]{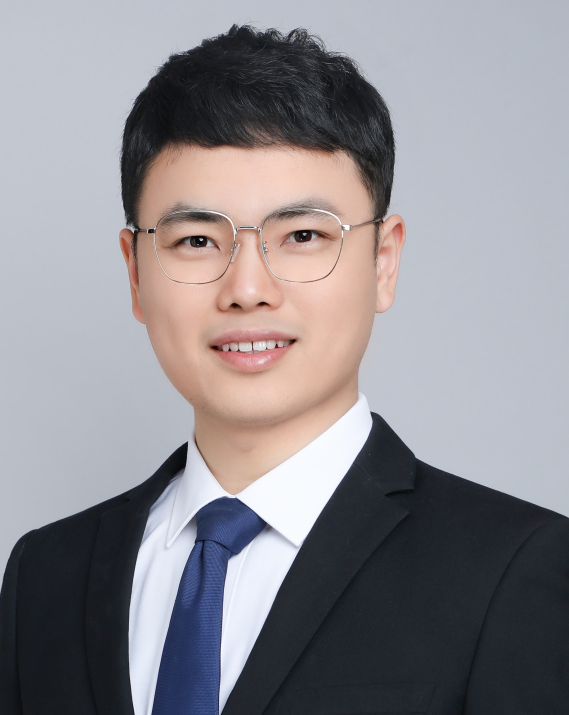}}]{xinghao chen}
received the BS and PhD degrees from the Department of Electronic Engineering, Tsinghua University, Beijing, China. He was a visiting PhD student with Imperial College London, U.K. He is currently with Huawei Noah’s Ark Lab. His research interests include deep learning and computer vision, with special interests in hand pose estimation, automated machine learning, and model compression.
\end{IEEEbiography}

\vspace{-12mm}

\begin{IEEEbiography}[{\includegraphics[width=1in,height=1.25in,clip,keepaspectratio]{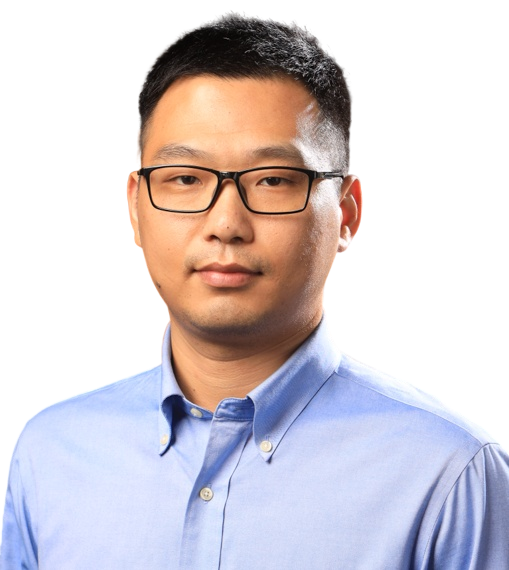}}]{Shaohui Lin}
received Ph.D. degree from Xiamen University, Xiamen, China, in 2019. He is currently a Research Professor at East China Normal University. He is the author of about 40 scientific articles at top venues, including IEEE TPAMI, TIP, CVPR, ICML, and AAAI, etc. He serves as the area chair for CVPR, IJCAI and PRCV, and reviewers for IEEE TPAMI, IJCV, TMM and CVPR, etc. His research interests include machine learning and computer vision.
\end{IEEEbiography}

\vspace{-12mm}

\begin{IEEEbiography}[{\includegraphics[width=1in,height=1.25in,clip,keepaspectratio]{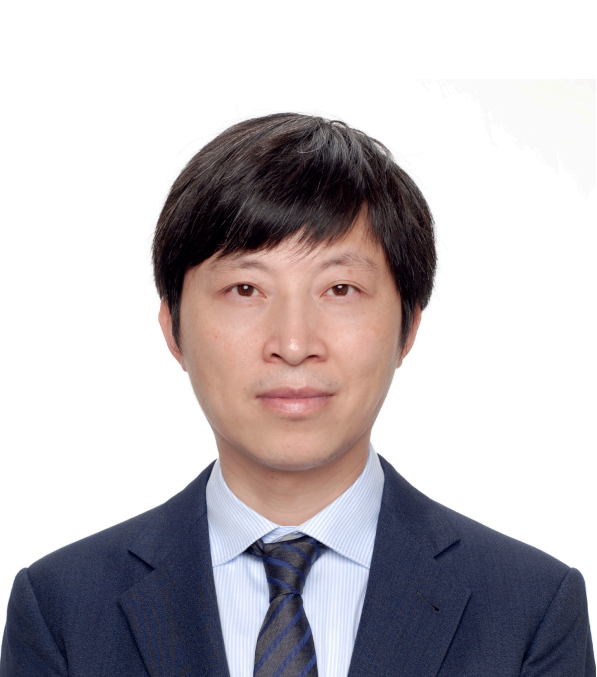}}]{Hongkai Xiong}
(Fellow, IEEE) is a Cheung Kong Professor and a Distinguished Professor at Shanghai Jiao Tong University (SJTU). Currently, he is with both Dept. Electronic Engineering and Dept. Computer Science and Engineering. From 2018-2020, he acted as the Vice Dean of Zhiyuan College in SJTU. He received the Ph.D. degree from Shanghai Jiao Tong University (SJTU), Shanghai, China, in 2003. Since then, he has been with the Department of Electronic Engineering, SJTU. He was an Associate Professor from 2005-2011 and an Assistant Professor from 2003-2005. From 2007 to 2008, he was a Research Scholar in the Department of Electrical and Computer Engineering, Carnegie Mellon University (CMU), Pittsburgh, PA, USA. During 2011-2012, he was a Scientist with the Division of Biomedical Informatics at the University of California (UCSD), San Diego, CA, USA. 

Dr. Xiong’s research interests include signal representation and wavelet analysis, image and video coding, multimedia communication and networking, computer vision, and machine learning. He published over 360 refereed journal and conference papers. He was the co-author of the TOP 1\% Paper Award of the 2022 ACM International Conference on Multimedia (ACM Multimedia’22), the Top 5\% Award at the 2024 IEEE International Conference on Image Processing (IEEE ICIP’24), the Top 10\% Paper Award at the 2016 IEEE Visual Communication and Image Processing (IEEE VCIP’16), the Best Student Paper Award at the 2014 IEEE Visual Communication and Image Processing (IEEE VCIP’14), the Best Paper Award at the 2013 IEEE International Symposium on Broadband Multimedia Systems and Broadcasting (IEEE BMSB’13), and the Top 10\% Paper Award at the 2011 IEEE International Workshop on Multimedia Signal Processing (IEEE MMSP’11).
%
%In 2022, Dr. Xiong was granted the First Prize of the Shanghai Science and Technology Progress Award. In 2021, Dr. Xiong was awarded the First Prize of Natural Science of the Chinese Institute of Electronics. In 2018, he was the recipient of the Shanghai Youth Science and Technology Distinguished Accomplishment Award. In 2017, he received both the Science and Technology Innovative Leader Talent Award, and Shanghai Academic Research Leader Talent Award. 
In 2014, he received both National Science Fund for Distinguished Young Scholar Award from Natural Science Foundation of China (NSFC) and Shanghai Youth Science and Technology Talent Award. %In 2013, he obtained Shanghai Shu Guang Scholar Award. In particular, he was given the recipients of Baosteel Excellent Faculty Award in both 2021 and
%2017, and the SJTU SMC-A Excellent Young Faculty Awards in both 2010
%and 2013. In both 2017 and 2011, he has been granted the First Prizes of the Shanghai Technology Innovation Awards. In 2009, he obtained the New Century Excellent Talent Award from Ministry of Education of China. 
From 2018-2021, he was the Associate Editor of IEEE Transactions on Circuits and Systems for Video Technology (TCSVT). He was elevated to a Fellow of the Chinese Institute of Electronics in 2023. He is also a Fellow of the IEEE.
\end{IEEEbiography}

\end{document}

%% file: table/main_compare.tex
\begin{table*}[t]
\captionsetup{font=small}
\small
\centering
\renewcommand\arraystretch{1.1}
\setlength\tabcolsep{4pt}
\caption{Quantitative comparison with state-of-the-art real-world SR methods on both synthetic and real-world benchmarks. Best and second best performance are highlighted in
\textcolor{red}{\textbf{red}} and \textcolor{blue}{\textbf{blue}}, respectively.}
\vspace{2mm}
\resizebox{\textwidth}{!}{
\begin{tabular}{@{}c|c|ccccccccccc@{}}
\toprule
Datasets                                                               & Metrics & StableSR \cite{wang2024exploiting} & DiffBIR \cite{lin2024diffbir} & ResShift \cite{yue2023resshift}  & SinSR \cite{wang2024sinsr} & SeeSR~\cite{wu2024seesr} & PASD~\cite{yang2024pixel} & Osediff~\cite{wu2024one} & VARSR~\cite{qu2025visual} & PURE~\cite{wei2025perceive}  & \textbf{RealSR-R1}  \\ \midrule
\multirow{9}{*}{\begin{tabular}[c]{@{}c@{}}\textit{OST-Val}\end{tabular}} & PSNR $\uparrow$   &20.51   &20.30   &\textcolor{red}{\textbf{20.98}}   &\textcolor{blue}{\textbf{20.73}}   &20.32   &20.70   &20.33   &20.68   &17.98 & 18.29 \\
    & SSIM $\uparrow$   &0.4847   &0.4550   &0.4930   &0.4747   &0.4818   &\textcolor{red}{\textbf{0.4921}}   &0.4760  &\textcolor{blue}{\textbf{0.4847}} &0.3816  &0.3963   \\
    & LPIPS $\downarrow$  &0.3825   &0.3966   &0.4304   &0.4053   &0.3569   &0.4671   &\textcolor{red}{\textbf{0.3482}}   &\textcolor{blue}{\textbf{0.3819}}  &0.4664 &0.4117   \\
    & DISTS $\downarrow$  &0.2159   &0.2189   &0.2546   &0.2305   &0.2134   &0.2429   &\textcolor{blue}{\textbf{0.2129}}   &0.2351  &0.2317 &\textcolor{red}{\textbf{0.2126}}   \\
    & FID $\downarrow$    &64.95   &67.18   &96.72   &81.54   &63.18   &81.31   &\textcolor{red}{\textbf{59.54}}   &72.71  &70.74 &\textcolor{blue}{\textbf{62.82}}   \\
    & NIQE $\downarrow$   &4.3539   &4.5481   &6.0288   &5.6849   &\textcolor{blue}{\textbf{4.2673}}   &5.2597   &4.2926   &5.5627  &4.9768 &\textcolor{red}{\textbf{4.1504}}   \\
    & MUSIQ $\uparrow$ &61.0076   &69.0289   &63.5278   &66.2410   &69.3768   &56.8370   &68.0248   &70.3423  &\textcolor{blue}{\textbf{70.5592}} &\textcolor{red}{\textbf{71.4819}}   \\
    & MANIQA $\uparrow$  &0.5886   &0.6485   &0.5666   &0.5591   &0.6527   &0.5244   &0.6455   &0.6336  &\textcolor{blue}{\textbf{0.6547}} &\textcolor{red}{\textbf{0.6858}}   \\
    & CLIPIQA $\uparrow$&0.5434   &0.7012   &0.6319   &0.6326   &0.6972   &0.4769   &0.6455   &0.7005 &\textcolor{blue}{\textbf{0.7038}} &\textcolor{red}{\textbf{0.7127}}    \\ 
    & TOPIQ $\uparrow$&0.6427   &0.7704   &0.6893   &0.7360   &0.7740   &0.5915   &0.7602   &0.7610  &\textcolor{blue}{\textbf{0.7775}} &\textcolor{red}{\textbf{0.8086}}   \\ \midrule
\multirow{9}{*}{\begin{tabular}[c]{@{}c@{}}\textit{DrealSR}\end{tabular}}     & PSNR $\uparrow$   & 28.03  & 26.71  & \textcolor{red}{\textbf{28.46}}  & \textcolor{blue}{\textbf{28.36}}  & 28.17  & 27.36  & 27.92  & 28.15 &24.39  &24.61    \\
    & SSIM $\uparrow$   & 0.7536  & 0.6571  & 0.7673  & 0.7515  & \textcolor{blue}{\textbf{0.7691}}  & 0.7073  & \textcolor{red}{\textbf{0.7835}}  &0.7654  & 0.6084  & 0.5797   \\
    & LPIPS $\downarrow$  & 0.3284  & 0.4557  & 0.4006  & 0.3665  & \textcolor{blue}{\textbf{0.3189}}  & 0.3760  & \textcolor{red}{\textbf{0.2968}} & 0.3540 & 0.4615  &0.4861   \\
    & DISTS $\downarrow$  & \textcolor{blue}{\textbf{0.2269}}  & 0.2748  & 0.2656  & 0.2485  & 0.2315  & 0.2531  & \textcolor{red}{\textbf{0.2165}} &0.2525  & 0.2756 & 0.2803  \\
    & FID  $\downarrow$   & 148.98  & 166.79  & 172.26  & 170.57  & \textcolor{blue}{\textbf{147.39}}  & 156.13  & \textcolor{red}{\textbf{135.30}}  & 157.59 &174.16  &171.56   \\
    & NIQE $\downarrow$   & 6.5239  & 6.3124  & 8.1249  & 6.9907  & 6.3967  & \textcolor{red}{\textbf{5.5474}}  & 6.4902 & 6.8709  & 6.3894  &\textcolor{blue}{\textbf{5.8207}}   \\
    & MUSIQ $\uparrow$ & 58.51  & 61.07  & 50.60  & 55.33 & 64.93  & 64.87 & 64.65 & \textcolor{red}{\textbf{68.0896}}  & 63.24  &\textcolor{blue}{\textbf{64.9527}}   \\
    & MANIQA $\uparrow$  & 0.5601  & 0.5930  & 0.4586  & 0.4884  & 0.6042  & 0.6169  & 0.5899 & \textcolor{blue}{\textbf{0.5961}}  & 0.5924  &\textcolor{red}{\textbf{0.6188}}   \\
    & CLIPIQA $\uparrow$& 0.6356  & 0.6395  & 0.5342  & 0.6383  & 0.6804  & 0.6808  & 0.6963 &\textcolor{blue}{\textbf{0.7196}} & 0.6789  &\textcolor{red}{\textbf{0.7241}}   \\ 
    & TOPIQ $\uparrow$&0.6208   &0.6899   &0.5646   &0.6211   &0.7026   &0.6398   &0.6758   &\textcolor{blue}{\textbf{0.7302}}  &0.7121  &\textcolor{red}{\textbf{0.7319}}    \\   \midrule
\multirow{9}{*}{\begin{tabular}[c]{@{}c@{}}\textit{RealSR}\end{tabular}}     & PSNR $\uparrow$   & 24.70  & 24.75  & \textcolor{red}{\textbf{26.31}}  &\textcolor{blue}{\textbf{26.28}}   & 25.18  & 25.21  & 25.15 &22.57 & 22.83 &22.89   \\
    & SSIM $\uparrow$   & 0.7085  & 0.6567  & \textcolor{red}{\textbf{0.7421}}  & \textcolor{blue}{\textbf{0.7347}}  & 0.7216  & 0.6798  & 0.7341 &0.7268 & 0.6079  &0.6146   \\
    & LPIPS $\downarrow$  & 0.3018  & 0.3636  & 0.3460  & 0.3188  & \textcolor{blue}{\textbf{0.3009}}  & 0.3380  & \textcolor{red}{\textbf{0.2921}} &0.3220  & 0.3821 &0.3871   \\
    & DISTS $\downarrow$  & 0.2288  & 0.2312  & 0.2498  & 0.2353  & \textcolor{blue}{\textbf{0.2223}}  & 0.2260  & \textcolor{red}{\textbf{0.2128}} &0.2356  & 0.2458 &0.2362  \\
    & FID $\downarrow$    & 128.51  & 128.99  & 141.71  & 135.93  & 125.55  & \textcolor{blue}{\textbf{124.29}}  & \textcolor{red}{\textbf{123.49}} &132.69  & 146.89 &128.85   \\
    & NIQE $\downarrow$   & 5.9122  & 5.5346  & 7.2635  & 6.2872  & \textcolor{blue}{\textbf{5.4081}}  & 5.4137  & 5.6476  & 6.0558 &5.8105  &\textcolor{red}{\textbf{4.9399}}   \\
    & MUSIQ $\uparrow$ & 65.78  & 64.98  & 58.43  & 60.80  & 69.77  & 68.75  & 69.09  & \textcolor{red}{\textbf{71.3023}} &66.7483  &\textcolor{blue}{\textbf{70.3638}}   \\
    & MANIQA $\uparrow$  & 0.6221  & 0.6246  & 0.5285  & 0.5385  & 0.6442  & 0.6487  & 0.6326  & \textcolor{red}{\textbf{0.6541}} &0.6310  &\textcolor{blue}{\textbf{0.6491}}   \\
    & CLIPIQA $\uparrow$& 0.6178  & 0.6463  & 0.5444  & 0.6122  & 0.6612  & 0.6620  & 0.6693  & \textcolor{blue}{\textbf{0.6981}} &0.6817  &\textcolor{red}{\textbf{0.7084}}   \\
    & TOPIQ $\uparrow$&0.6809   &0.7313   &0.6242   &0.6411   &0.7354   &0.5988   &0.7106   &\textcolor{red}{\textbf{0.7422}}  &0.7165  &\textcolor{blue}{\textbf{0.7371}}   \\ \midrule
\multirow{4}{*}{\begin{tabular}[c]{@{}c@{}}\textit{RealLQ250}\end{tabular}}  & NIQE $\downarrow$   &4.5282   &4.7524   & 5.3275  &5.1777   &4.4451   &\textcolor{blue}{\textbf{4.3240}} &4.4181  &5.2372  &4.7158 & \textcolor{red}{\textbf{4.0454}}  \\
    & MANIQA $\uparrow$ &67.3269   &71.7762   &65.7151   &66.8540   &70.3724   &65.8979   &70.9796  &\textcolor{red}{\textbf{73.8639}} &72.6105 &\textcolor{blue}{\textbf{73.4231}}   \\
    & MUSIQ $\uparrow$  &0.6352   &\textcolor{red}{\textbf{0.6686}}   &0.5753   &0.5825   & 0.5927  &0.6142   &0.6478  &\textcolor{blue}{\textbf{0.6704}} &0.6886 &0.6601  \\
    & CLIPIQA $\uparrow$&0.6566   &0.7667   &0.6336   &0.6781   &0.7062   &0.6183   &0.7097  &0.7556 &\textcolor{blue}{\textbf{0.7753}}  &\textcolor{red}{\textbf{0.7813}} \\ 
    & TOPIQ $\uparrow$&0.6893   &0.7629   &0.6795   &0.7084   &0.6939   &0.6572   &0.7286   &0.7653 &\textcolor{blue}{\textbf{0.7781}}  &\textcolor{red}{\textbf{0.7884}} \\ \bottomrule
\end{tabular}
}
\label{tab:methods}
\vspace{-0.5cm}
\end{table*}

%% file: table/rebuttal_time.tex
\begin{table}[h]
\centering
\centering

\caption{model complexity comparisons with the output size of $512 \times 512$.}
\label{tab:re_time}
\resizebox{\columnwidth}{!}{
\begin{tabular}{lcccccc}
\hline
Method    & Time(s)  & \multicolumn{1}{c}{PSNR↑} & \multicolumn{1}{c}{SSIM↑}  & \multicolumn{1}{c}{LPIPS↓}  & \multicolumn{1}{c}{DISTS↓}   & \multicolumn{1}{c}{FID↓}   \\ \hline
PURE      & 316      & 17.98                     & 0.3816                     & 0.4664                      & 0.2317                       & 70.74                      \\
RealSR-R1 & 372      & \textbf{18.29}            & \textbf{0.3963}            & \textbf{0.4117}             & \textbf{0.2126}              & \textbf{62.82}             \\ \hline
          & Param(M) & \multicolumn{1}{c}{NIQE↓} & \multicolumn{1}{c}{MUSIQ↑} & \multicolumn{1}{c}{MANIQA↑} & \multicolumn{1}{c}{CLIPIQA↑} & \multicolumn{1}{c}{TOPIQ↑} \\ \hline
PURE      & 7080     & 4.9768                    & 70.5592                    & 0.6547                      & 0.7038                       & 0.7775                     \\
RealSR-R1 & 7080     & \textbf{4.1504}           & \textbf{71.4819}           & \textbf{0.6858}             & \textbf{0.7127}              & \textbf{0.8086}            \\ \hline
\end{tabular}}
\end{table}

%% file: table/rebuttal_rain.tex
\begin{table}[h]
    \centering
        \centering

        \caption{Result on image deraining task.}
        \label{tab:re_rain}
        \resizebox{\columnwidth}{!}{
        \begin{tabular}{l|cccccc}
        \toprule
                                 & \multicolumn{5}{c}{Rian100H}                                            \\
        \multirow{-2}{*}{Method} & PSNR$\uparrow$ & LPIPS$\downarrow$ & MUSIQ$\uparrow$ & MANIQA$\uparrow$  & CLIPIQA$\uparrow$ &  TOPIQ$\uparrow$     \\ 
        \midrule
        MPRNet~\cite{zamir2021multi}                  &30.41       &0.158        &69.5733         &0.6788                 &0.6534 &0.6676                               \\
        IR-SDE~\cite{luo2023image}                  &31.65       &0.047       &70.6452         &0.6811               & 0.7782 &0.7085                              \\
         RealSR-R1                  &\textbf{33.59}       &\textbf{0.419}        &\textbf{72.3578}         &\textbf{0.7182}                 &\textbf{0.8196}    &\textbf{0.7741}                           \\
        \bottomrule
        \end{tabular}}

\end{table}

%% file: table/aba_mian.tex
\begin{table*}[t]
    \centering
    \begin{minipage}[t]{0.475\linewidth}
        \centering
        %\vspace{-2mm}
        \caption{Ablation of the VLCot on OST-Val.}
        \label{tab:CoT}
        \resizebox{\textwidth}{!}{
        \begin{tabular}{l|lllll}
        \toprule
                                 & \multicolumn{5}{c}{OST-Val}                                            \\
        \multirow{-2}{*}{Method} & SSIM$\uparrow$ & LPIPS$\downarrow$ & MANIQA$\uparrow$ & CLIPIQA$\uparrow$ &  TOPIQ$\uparrow$     \\ 
        \midrule
        RealSR-R1                  &\textbf{0.3963}       &\textbf{0.4117}        &\textbf{0.6858}         &\textbf{0.7127}                 &\textbf{0.8086}                               \\
        -VLCoT GRPO                  &0.3792       & 0.4481       &0.6735         &0.7096                & 0.7859                              \\
        -Degradation output      &0.3564       & 0.4618       & 0.6651        & 0.6993             &  0.7876                             \\
        -Understanding output      & 0.3512      & 0.4593       &  0.6408           & 0.6819         &  0.7784                             \\
        -Multi output            & 0.3359      &  0.4875      &  0.6267       &  0.6715              &   0.7547                            \\ 
        \bottomrule
        \end{tabular}}
    \end{minipage}%
    \hspace{0.5cm}
    \begin{minipage}[t]{0.465\linewidth}
        \centering
        \scriptsize
        \vspace{-1mm}
        \caption{Ablation on reward functions.}
        \label{tab:reward}
        \resizebox{\textwidth}{!}{
        \begin{tabular}{l|lllll}
        \toprule
                                 & \multicolumn{5}{c}{OST-Val}                                            \\
        \multirow{-2}{*}{Reward} & SSIM$\uparrow$ & LPIPS$\downarrow$ & MANIQA$\uparrow$ & CLIPIQA$\uparrow$ &  TOPIQ$\uparrow $    \\ 
        \midrule
        RealSR-R1                  &\textbf{0.3963}       &\textbf{0.4117}        & \textbf{0.6858}        &  \textbf{0.7127}               & \textbf{0.8086}                              \\
        -Understanding                  & 0.3874       &0.4353        & 0.6804        &0.7089                & 0.8024                              \\
        -Degradation      &0.3765       &0.4649        &0.6833         &0.7101              &0.8015                               \\
        -Generation       &0.3816       &0.4432        & 0.6798            &0.7055          & 0.7831                              \\
        \bottomrule
        \end{tabular}}
    \end{minipage}
    \vspace{-4mm}
\end{table*}

%% file: table/aba_reward_U.tex
\begin{table}[h]
    \centering
        \centering

        \caption{Ablation of tag extraction strategies within the Understanding reward.}
        \label{tab:U_reward}
        \resizebox{\columnwidth}{!}{
        \begin{tabular}{l|ccccc}
        \toprule
                                 & \multicolumn{5}{c}{OST-Val}                                            \\
        \multirow{-2}{*}{Method} & SSIM$\uparrow$ & LPIPS$\downarrow$ & MANIQA$\uparrow$ & CLIPIQA$\uparrow$ &  TOPIQ$\uparrow$     \\ 
        \midrule
        RAM                  &0.3847       &0.4226        &0.6759         &0.7065                 &0.8019                               \\
        Qwen2.5-VL                  &\textbf{0.4023}       & 0.4136       &0.6813         &0.7082               & 0.8032                              \\
         Ground-SAM                  &0.3963       &\textbf{0.4117}        &\textbf{0.6858}         &\textbf{0.7127}                 &\textbf{0.8086}                               \\
        \bottomrule
        \end{tabular}}

\end{table}

%% file: table/reward_G.tex
\begin{table}[t]
    \centering
        \centering
        \scriptsize
        \caption{Ablation of the generation reward. The perceptual metric is computed as a weighted combination of PSNR, SSIM, LPIPS, and DISTS scores, while the generative quality metric is defined as a weighted aggregation of CLIPIQA, MANIQA, MUSIQ, and TOPIQ scores.}
        \label{tab:G_reward}
        \resizebox{\columnwidth}{!}{
        \begin{tabular}{l|cccccc}
        \toprule
                                 & \multicolumn{5}{c}{OST-Val}                                            \\
        \multirow{-2}{*}{Method} & PSNR$\uparrow$ & SSIM$\uparrow$ & LPIPS$\downarrow$ & MANIQA$\uparrow$ & CLIPIQA$\uparrow$ &  TOPIQ$\uparrow$     \\ 
        \midrule
        Perceptual metrics       &\textbf{19.89}           &\textbf{0.4298}       &\textbf{0.3672}        &0.6578         &0.6849                 &0.7633                               \\
        Generate metrics         &17.73        &0.3595       & 0.4781       &\textbf{0.6869}         &0.7106                & \textbf{0.8168}                              \\
        Qwen2.5-VL       &18.29           &0.3963       &0.4117        &0.6858         &\textbf{0.7127}                 &0.8086                               \\
        \bottomrule
        \end{tabular}}
\end{table}

%% file: table/rebuttal_step.tex
\begin{table}[h]
\centering
\centering
\scriptsize

\caption{Ablation of different step of RealSR-R1.}
\label{tab:re_step}
\begin{tabular}{lccccc}
\hline
Method     & \multicolumn{1}{c}{SSIM↑} & \multicolumn{1}{c}{LPIPS↓} & \multicolumn{1}{c}{MANIQA↑} & \multicolumn{1}{c}{CLIPIQA↑} & \multicolumn{1}{c}{TOPIQ↑} \\ \hline
Two-step   & 0.3924                    & 0.4248                     & 0.6781                      & 0.7106                       & 0.7913                     \\
Three-step & \textbf{0.3963}           & \textbf{0.4117}            & \textbf{0.6858}             & \textbf{0.7127}              & \textbf{0.8086}            \\ \hline
\end{tabular}
\end{table}